%% file: main.tex
\documentclass[10pt,twocolumn,letterpaper]{article}

\usepackage[pagenumbers]{cvpr} 

\usepackage{graphicx}
\usepackage{amsmath}
\usepackage{amssymb}
\usepackage{booktabs}
\usepackage{tabularx}
\usepackage{multirow}
\usepackage{float}
\usepackage{caption}
\usepackage{subcaption}
\usepackage{xcolor}
\usepackage{enumitem}
\usepackage[frozencache=true,cachedir=_minted-main]{minted}
\usepackage{appendix}

\usepackage{mathtools}
\usepackage[nolist]{acronym}

\definecolor{bg}{HTML}{111111}

\usepackage[pagebackref,breaklinks,colorlinks]{hyperref}

\usepackage[capitalize]{cleveref}
\crefname{section}{Sec.}{Secs.}
\Crefname{section}{Section}{Sections}
\Crefname{table}{Table}{Tables}
\crefname{table}{Tab.}{Tabs.}

\def\cvprPaperID{6413} 
\def\confName{CVPR}
\def\confYear{2023}

\newcommand{\name}{\emph{BiasBed}}

\begin{document}

\begin{acronym}
\acro{cnn}[CNN]{convolutional neural network}
\acro{ood}[OOD]{out-of-domain}
\end{acronym}
\title{BiasBed -- Rigorous Texture Bias Evaluation}

\author{
Nikolai Kalischek\textsuperscript{1}$\qquad$
Rodrigo Caye Daudt\textsuperscript{1}$\qquad$
Torben Peters\textsuperscript{1}\\
Reinhard Furrer\textsuperscript{2}$\quad$
Jan D.\ Wegner\textsuperscript{3}$\quad$
Konrad Schindler\textsuperscript{1}\\
\textsuperscript{1}Photogrammetry and Remote Sensing, ETH Zürich\\
\textsuperscript{2}Institute for Mathematics, University of Zurich\\
\textsuperscript{3}Institute for Computational Science, University of Zurich
}

\maketitle


\begin{abstract}

The well-documented presence of texture bias in modern convolutional neural networks has led to a plethora of algorithms that promote an emphasis on shape cues, often to support generalization to new domains. Yet, common datasets, benchmarks and general model selection strategies are missing, and there is no agreed, rigorous evaluation protocol. In this paper, we investigate difficulties and limitations when training networks with reduced texture bias. In particular, we also show that proper evaluation and meaningful comparisons between methods are not trivial. We introduce \name{}, a testbed for texture- and style-biased training, including multiple datasets and a range of existing algorithms. It comes with an extensive evaluation protocol that includes rigorous hypothesis testing to gauge the significance of the results, despite the considerable training instability of some style bias methods. Our extensive experiments, shed new light on the need for careful, statistically founded evaluation protocols for style bias (and beyond). \Eg, we find that some algorithms proposed in the literature do not significantly mitigate the impact of style bias at all. With the release of \name{}, we hope to foster a common understanding of consistent and meaningful comparisons, and consequently faster progress towards learning methods free of texture bias. Code is available at \url{https://github.com/D1noFuzi/BiasBed}

\end{abstract}

\section{Introduction} \label{sec:intro}
Visual object recognition is fundamental to our daily lives. Identifying and categorizing objects in the environment is essential for human survival, indeed our brain is able to assign the correct object class within a fraction of a second, independent of substantial variations in appearance, illumination and occlusion \cite{logothetis1996visual}. Recognition mainly takes place along the ventral pathway \cite{dicarlo2012does}, i.e., visual perception induces a hierarchical processing stream from local patterns to complex features, in feed-forward fashion \cite{logothetis1996visual}. Signals are filtered to low frequency and used in parallel also in a top-down manner\cite{bar2006top}, emphasizing the robustness and invariance to deviations in appearance. 

Inspired by our brain's visual perception, convolutional neural architectures build upon the hierarchical intuition, stacking multiple convolutional layers to induce feed-forward learning of basic concepts to compositions of complex objects. Indeed, early findings suggested that neurons in deeper layers are activated by increasingly complex shapes, while the first layers are mainly tailored towards low-level features such as color, texture and basic geometry. However, recent work indicates the opposite: \acp{cnn} exhibit a strong bias to base their decision on texture cues \cite{geirhos2021partial, gatys2017texture, Hermann2020Advances, geirhos2018imagenet}, which heavily influences their performance, in particular under domain shifts, which typically affect local texture more than global shape.

This inherent texture bias has led to a considerable body of work that tries to minimize texture and style bias and shift towards ``human-like" shape bias in object recognition. Common to all approaches is the principle of incorporating adversarial texture cues into the learning pipeline -- either directly in input space \cite{geirhos2018imagenet, hendrycks2021many, li2020shape} or implicitly in feature space \cite{nuriel2021permuted, nam2021reducing, kang2022style, jeon2021feature}. Perturbing the texture cues in the input forces the neural network to make ``texture-free" decisions, thus focusing on the objects' shapes that remain stable during training. While texture cues certainly boost performance in fine-grained classification tasks, where local texture patterns may indicate different classes, they can cause serious harm when the test data exhibit a domain shift \wrt training. In this light, texture bias can be seen as a main reason for degrading domain generalization performance, and various algorithms have been developed to improve generalization to new domains with different texture properties (respectively image styles), \eg,~\cite{geirhos2018imagenet, hendrycks2021many, li2020shape,nuriel2021permuted, nam2021reducing, kang2022style, jeon2021feature, wang2022feature}. 

However, while a considerable number of algorithms have been proposed to address texture bias, they are neither evaluated on common datasets nor with common evaluation metrics. Moreover they often employ inconsistent model selection criteria or do not report at all how model selection is done. 
With the present paper we promote the view that:
\begin{itemize}[parsep=2pt]
    \item the large domain shifts induced by texture-biased training cause large fluctuations in accuracy, which call for a  particularly rigorous evaluation;
    \item model selection has so far been ignored in the literature; together with the high training volatility, this may have lead to overly optimistic conclusions from spurious results;
    \item in light of the difficulties of operating under drastic domain shifts, experimental results should include a notion of uncertainty in the evaluation metrics.
\end{itemize}

Motivated by these findings, we have implemented \name{}, an open-source PyTorch \cite{paszke2019pytorch} testbed that comes with six datasets of different texture and shape biases, four adversarial robustness datasets and eight fully implemented algorithms. Our framework allows the addition of new algorithms and datasets with few lines of code, including full flexibility of all parameter settings. In order to tackle the previously discussed limitations and difficulties for evaluating such algorithms, we have added a carefully designed evaluation pipeline that includes training multiple runs and employing multiple model selection methods, and we report all results based on sound statistical hypothesis tests -- all run with a single command. 
In addition to our novel framework, the present paper makes the following contributions:
\begin{itemize}[parsep=2pt]
    \item We highlight shortcomings in the evaluation protocols used in recent work on style bias, including the observation that there is a very high variance in the performance of different runs of the same algorithm, and even between different checkpoints in the same run with similar validation scores.
    \item We develop and openly release a testbed that rigorously compares different algorithms using well-established hypothesis testing methods. This testbed includes several of the most prominent algorithms and datasets in the field, and is easily extensible.
    \item We observe in our results that current algorithms on texture-bias datasets fail to surpass simple ERM in a statistically significant way, which is the main motivation for this work and for using rigorous hypothesis tests for evaluating style bias algorithms.
\end{itemize}

In Sec.~\ref{sec:relatedwork}, we provide a comprehensive overview of existing work on reducing texture bias. Furthermore we describe the main forms of statistical hypothesis testing. We continue in Sec.~\ref{sec:biasedlearning} with a formal introduction to biased learning, and systematically group existing algorithms into families with common properties. In Sec.~\ref{sec:evaluationpractices}, we investigate previous evaluation practices in detail, discuss their limitations, and give a formal introduction to hypothesis testing. Sec.~\ref{sec:BiasBedframework} describes our proposed \name{} evaluation framework in detail, while Sec.~\ref{sec:experiments} presents experiments, followed by a discussion (Sec.~\ref{sec:discussion}), conclusions (Sec.~\ref{sec:conclusion}) and a view towards the broader impact of our work (Sec.~\ref{sec:broaderimpact}).  

\section{Related work}\label{sec:relatedwork}

\begin{figure*}[th]
    \setlength{\tabcolsep}{1pt}
    \renewcommand{\arraystretch}{0.6}
    \centering
    \begin{subfigure}[b]{0.24\textwidth}
        \begin{tabular}{cc}
            \includegraphics[width=0.45\textwidth, height=0.45\textwidth]{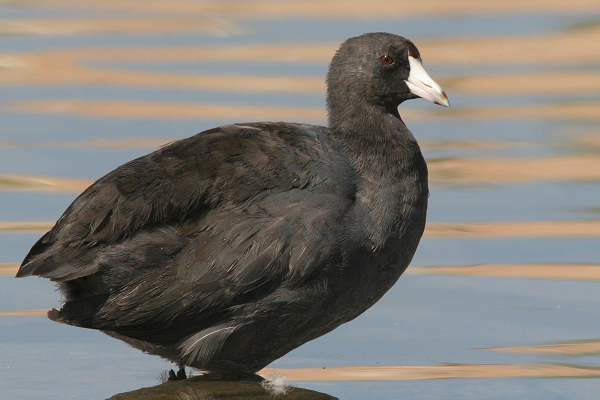} &
            \includegraphics[width=0.45\textwidth, height=0.45\textwidth]{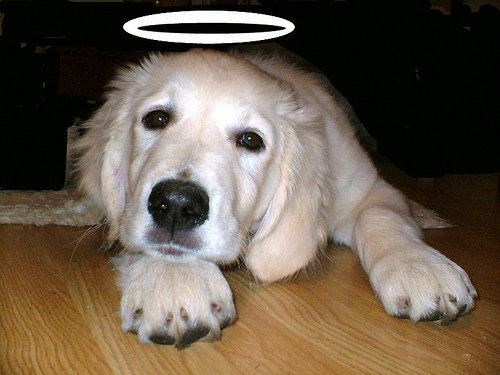} \\
            \includegraphics[width=0.45\textwidth, height=0.45\textwidth]{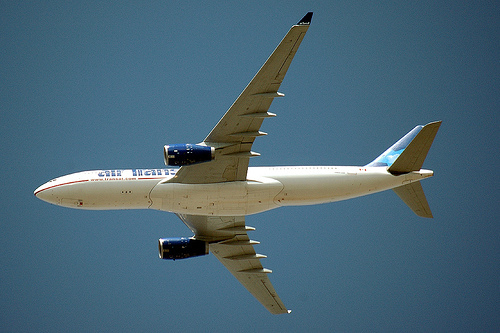} &
            \includegraphics[width=0.45\textwidth, height=0.45\textwidth]{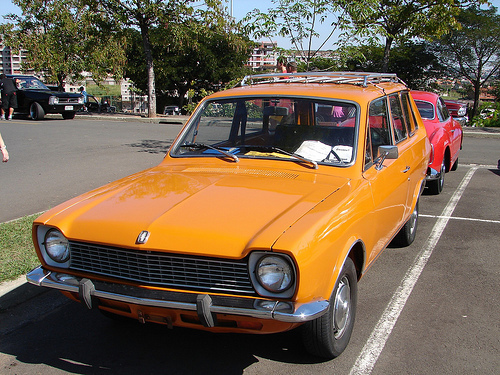}
        \end{tabular}
        \subcaption{ImageNet}
    \end{subfigure}
    \begin{subfigure}[b]{0.24\textwidth}
        \begin{tabular}{cc}
            \includegraphics[width=0.45\textwidth]{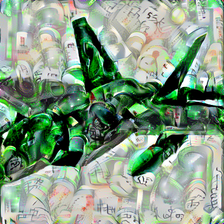} &
            \includegraphics[width=0.45\textwidth]{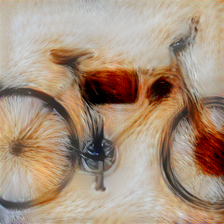} \\
            \includegraphics[width=0.45\textwidth]{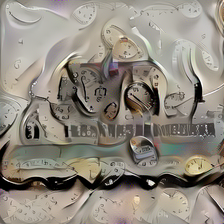} &
            \includegraphics[width=0.45\textwidth]{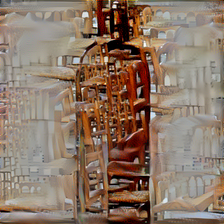}
        \end{tabular}
        \subcaption{Cue-Conflict}
    \end{subfigure}
    \begin{subfigure}[b]{0.24\textwidth}
        \begin{tabular}{cc}
            \includegraphics[width=0.45\textwidth]{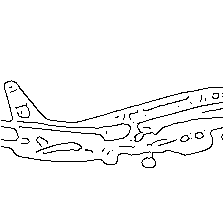} &
            \includegraphics[width=0.45\textwidth]{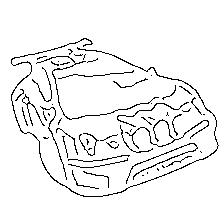} \\
            \includegraphics[width=0.45\textwidth]{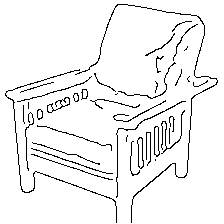} &
            \includegraphics[width=0.45\textwidth]{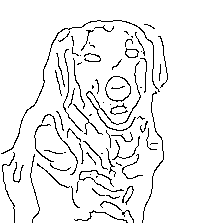}
        \end{tabular}
        \subcaption{Edge}
    \end{subfigure}
    \begin{subfigure}[b]{0.24\textwidth}
        \begin{tabular}{cc}
            \includegraphics[width=0.45\textwidth]{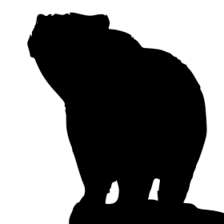} &
            \includegraphics[width=0.45\textwidth]{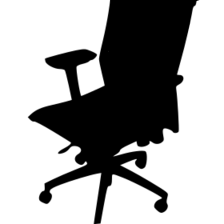} \\
            \includegraphics[width=0.45\textwidth]{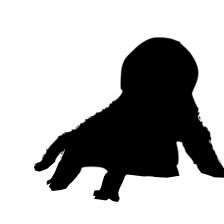} &
            \includegraphics[width=0.45\textwidth]{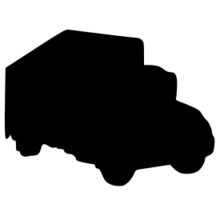}
        \end{tabular}
        \subcaption{Silhouette}
    \end{subfigure}
    \begin{subfigure}[b]{0.24\textwidth}
        \begin{tabular}{cc}
            \includegraphics[width=0.45\textwidth]{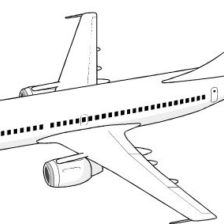} &
            \includegraphics[width=0.45\textwidth]{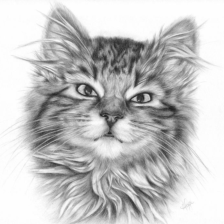} \\
            \includegraphics[width=0.45\textwidth]{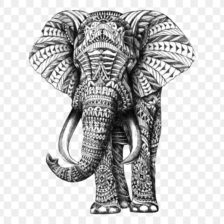} &
            \includegraphics[width=0.45\textwidth]{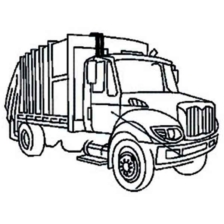}
        \end{tabular}
        \subcaption{Sketch}
    \end{subfigure}
    \begin{subfigure}[b]{0.24\textwidth}
        \begin{tabular}{cc}
            \includegraphics[width=0.45\textwidth]{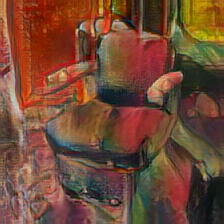} &
            \includegraphics[width=0.45\textwidth]{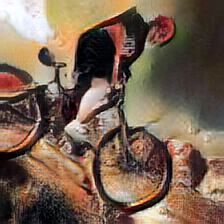} \\
            \includegraphics[width=0.45\textwidth]{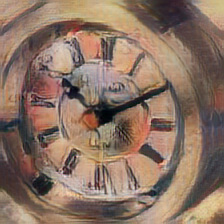} &
            \includegraphics[width=0.45\textwidth]{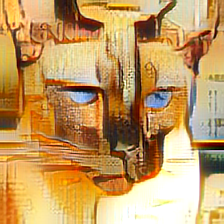}
        \end{tabular}
        \subcaption{Stylized ImageNet}
    \end{subfigure}
    \begin{subfigure}[b]{0.24\textwidth}
        \begin{tabular}{cc}
            \includegraphics[width=0.45\textwidth]{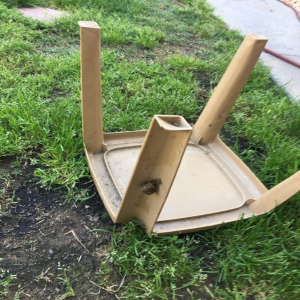} &
            \includegraphics[width=0.45\textwidth]{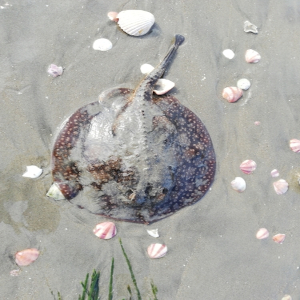} \\
            \includegraphics[width=0.45\textwidth]{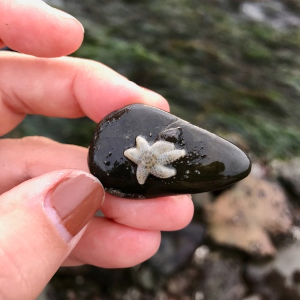} &
            \includegraphics[width=0.45\textwidth]{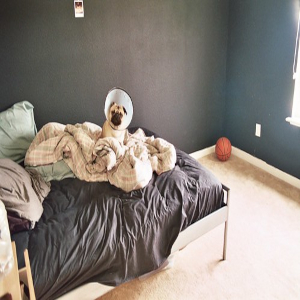}
        \end{tabular}
        \subcaption{ImageNet-A}
    \end{subfigure}
    \begin{subfigure}[b]{0.24\textwidth}
        \begin{tabular}{cc}
            \includegraphics[width=0.45\textwidth]{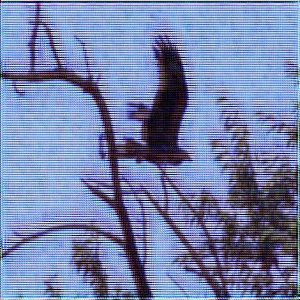} &
            \includegraphics[width=0.45\textwidth]{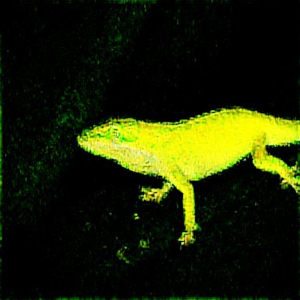} \\
            \includegraphics[width=0.45\textwidth]{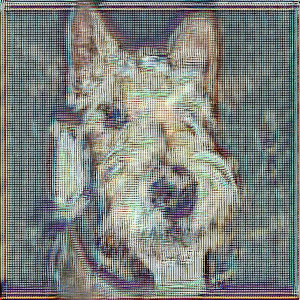} &
            \includegraphics[width=0.45\textwidth]{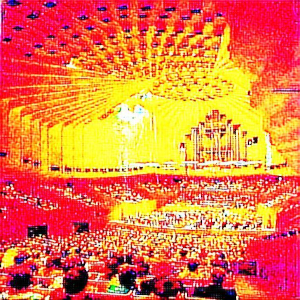}
        \end{tabular}
        \subcaption{DeepAug EDSR and CAE}
    \end{subfigure}
     
    \caption{Samples of datasets used in our experiments. These datasets (a-f) aim to capture different facets of what is informally known as ``texture bias". They attempt to decouple large scale features, such as image structure and object shapes, from small scale texture features. While humans can easily recognize the objects in these images in most cases, convolutional neural networks often struggle to do so with the same accuracy that they achieve with natural images; samples from (g) and (h) can be used to test for adversarial robustness.}
    \label{fig:datasets}
\end{figure*}

\paragraph{Texture and style bias.}

The seminal work of Geirhos \etal \cite{geirhos2018imagenet} showed that modern neural networks are heavily biased towards local patterns, commonly referred to as texture or style.\footnote{In this paper we prefer the word ``texture", but interchangeably use ``style" as part of already set names and expressions, as in ``style transfer".} The finding that, contrary to earlier hypotheses \cite{lecun2015deep, kriegeskorte2015deep}, the network output is dominated by domain-specific texture rather than generic, large-scale shape cues helps to explain their poor ability to generalize to unseen image domains. As a consequence, several authors have tried to address the issue under the umbrella term \emph{Domain Generalization} \cite{nam2021reducing, kang2022style, jeon2021feature}, although \emph{texture debiasing} would be a more accurate description of their approach: they take advantage of neural style transfer ideas \cite{kalischek2021light, gatys2017texture} and alter the texture features, thereby forcing the model to rely more on shape features. To that end, Geirhos \etal \cite{geirhos2018imagenet} put forward a new \textit{stylized} ImageNet dataset. That dataset contains images that are blended with a random texture image to generate a diverse set of ``texture-independent" samples, such that the contained texture is not informative about the class label anymore. However, enforcing shape bias tends to deteriorate performance on the source domain, \ie the domain trained on, \eg ImageNet. To balance in-domain and \ac{ood} performance, Li \etal \cite{li2020shape} propose debiased learning by linearly interpolating the target labels between the texture class and the content class.

Inspired by neural style transfer, a whole palette of work \cite{nuriel2021permuted, nam2021reducing, kang2022style, jeon2021feature, wang2022feature} deals with changing the texture statistics in feature space. The authors of \cite{nuriel2021permuted} randomly swap mean and standard deviation along the spatial dimension in certain network layers, while \cite{wang2022feature} add learnable encoders and noise variables for each statistic to align them over different layers. In \cite{nam2021reducing} an additional adversarial content randomization block tricks the network into predicting the right style despite changed content. Jeon \etal \cite{jeon2021feature} sample new style vectors by defining Gaussians over the channel dimension. 

\paragraph{Evaluation frameworks.}

Comparing algorithms based on some relevant quantitative performance metric has become the norm in the deep learning community. A common strategy for domain generalization, which by definition requires two or more different datasets, is to evaluate each algorithm on the held-out testing portion of every dataset. From the outcomes it is typically concluded that methods which, on average, perform better is superior. To account for variance in learning procedures \cite{bouthillier2021accounting}, a test bed similar to ours \cite{gulrajani2020search} trains multiple models per algorithm and uses a mean of means $\mu$ over different datasets as the final metric, i.e., method A is considered better than method B if $\mu_A > \mu_B$. Clearly, this conclusion is not necessarily justified from a statistical viewpoint \cite{dror2017replicability}, as it does not distinguish between true impact and random effects \cite{bouthillier2021accounting}. 

In fact, statistical testing offers a formal theory for comparing multiple algorithms over multiple datasets \cite{neyman1928use}, a tool that is routinely used in scientific fields like physics or medicine. Hypothesis tests are employed to answer the question whether two (or more) algorithms differ significantly, given performance estimates on one or more datasets. Depending on assumptions about the scores, the comparison may require a parametric or a non-parametric test. In machine learning (ML) the requirements for parametric tests like the repeated-measures ANOVA \cite{fisher1992statistical} are typically not met, \eg, scores may not be normally and spherically distributed \cite{demvsar2006statistical}.
Non-parametric tests like the Friedman test \cite{friedman1937use} require fewer assumptions and are  more generally applicable. A recent line of work \cite{wainer2022bayesian} proposed a Bayesian form of the Bradley-Terry model to determine how confident one can be that one algorithm is better than another. 

\section{Biased learning}\label{sec:biasedlearning}

Shape-biased learning tries to minimize the error that stems from distribution shifts between the training and test domains. Formally, we train a neural network $g$ on a training dataset \{$(x_i,y_i)\}_{i\in [n]}$, where the value $x_i \in \mathcal{X}$ of a random variable $X$ is the input to the network and $y_i \in \mathcal{Y}$ of a random variable $Y$ is the corresponding ground truth. Let $P(X)$ be the source domain distribution. In shape-biased learning however, we test on samples $x \in \mathcal{X}'$ such that $P(X) \neq P(X')$. In our setting the distribution shift mainly results from changes in texture, \eg going from photos to cartoons or sketches. Shape-biased learning is a special case of domain generalization \cite{li2017deeper}, which also encompasses distribution shifts due to factors other than texture. 

Existing literature can be grouped into two families, input-augmented \cite{geirhos2018imagenet, hendrycks2021many, li2020shape} and feature-augmented \cite{nuriel2021permuted, nam2021reducing, kang2022style, jeon2021feature, wang2022feature} methods. The former create style-randomized versions of the training data, whereas the latter apply stylization to the latent representations inside a neural network. The methodology to transfer style remains the same for both. Given two images $I, I' \in \mathbb{R}^{H\times W\times 3}$ and a pretrained network (\eg, VGG-19 \cite{simonyan2014very}), feature statistics at suitable network layers are computed for both images. The most common approach is to use first and second moments as in \cite{huang2017arbitrary}. Let $F_l(x) \in \mathbb{R}^{H\times W\times C_l}$ be the $l$-th feature map with input $x$, $\mu_{F_l(x)} \in \mathbb{R}^{C_l}$ its channel-wise mean (averaged across the spatial dimensions) and $\sigma_{F_l(x)} \in \mathbb{R}^{C_l}$ the corresponding standard deviation. Then style transfer boils down to \cite{huang2017arbitrary} 
\begin{equation}
    F_l^{\text{new}}(x) = \sigma_{F_l(x')} \bigg(\frac{x - \mu_{F_l(x)}}{\sigma_{F_l(x)}}\bigg) + \mu_{F_l(x')},
\end{equation}
where $x, x'$ are two different samples.
In case of input augmentation, the (stylized) encodings are propagated through a decoder to generate images, either offline \cite{geirhos2018imagenet} or on the fly \cite{li2020shape}. This process is often applied to features at various representation levels to capture textures at different scales. Feature augmentation methods swap the statistics within their classification network and directly output the corresponding class predictions.

\section{Evaluation practices}\label{sec:evaluationpractices}

We first give an overview of current evaluation practices and point out limitations when moving out of domain. Then follows an introduction to formal hypothesis testing.

\begin{figure*}[h!tb]
     \centering
     \begin{subfigure}[b]{0.33\textwidth}
         \centering
         \includegraphics[width=\textwidth]{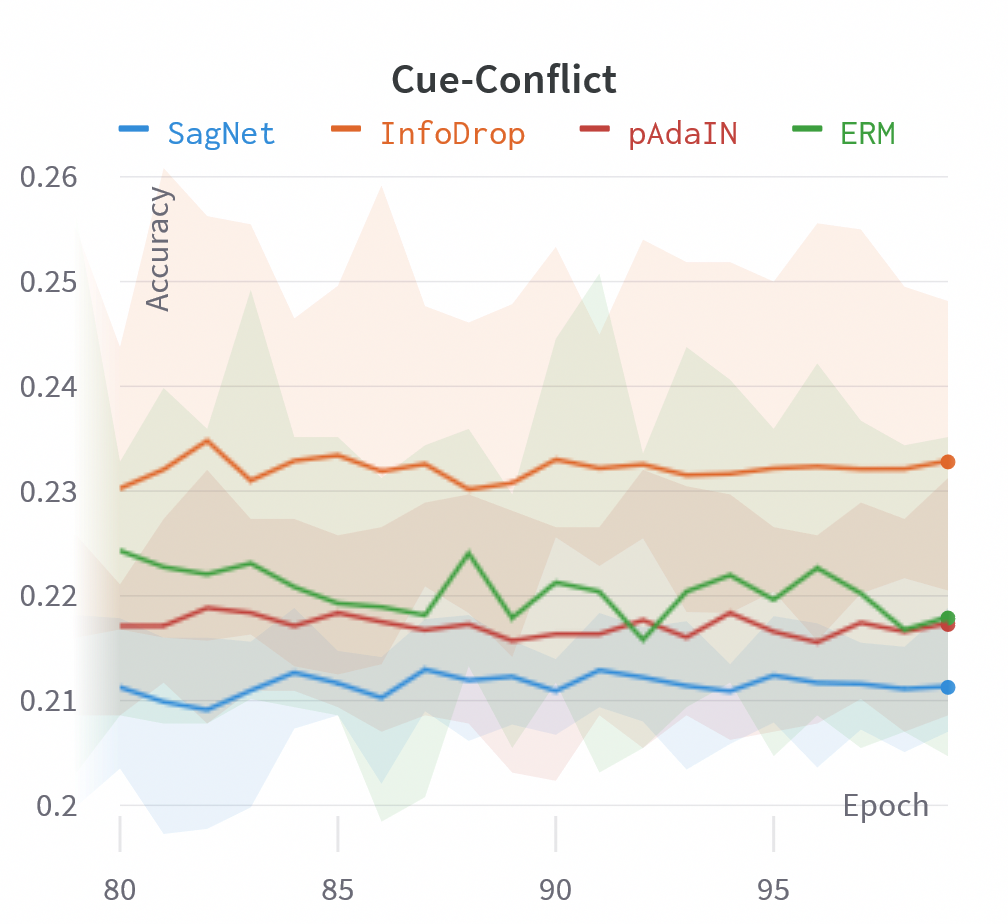}
     \end{subfigure}
     \hfill
     \begin{subfigure}[b]{0.33\textwidth}
         \centering
         \includegraphics[width=\textwidth]{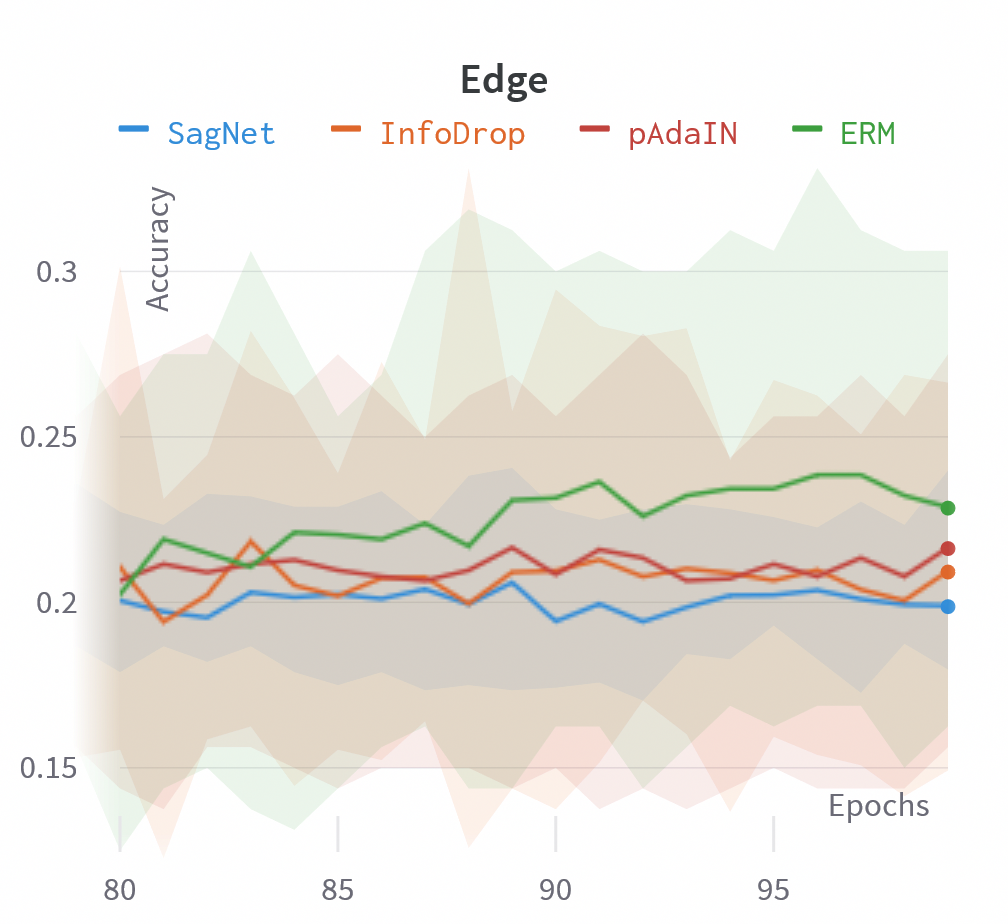}
     \end{subfigure}
     \hfill
     \begin{subfigure}[b]{0.33\textwidth}
         \centering
         \includegraphics[width=\textwidth]{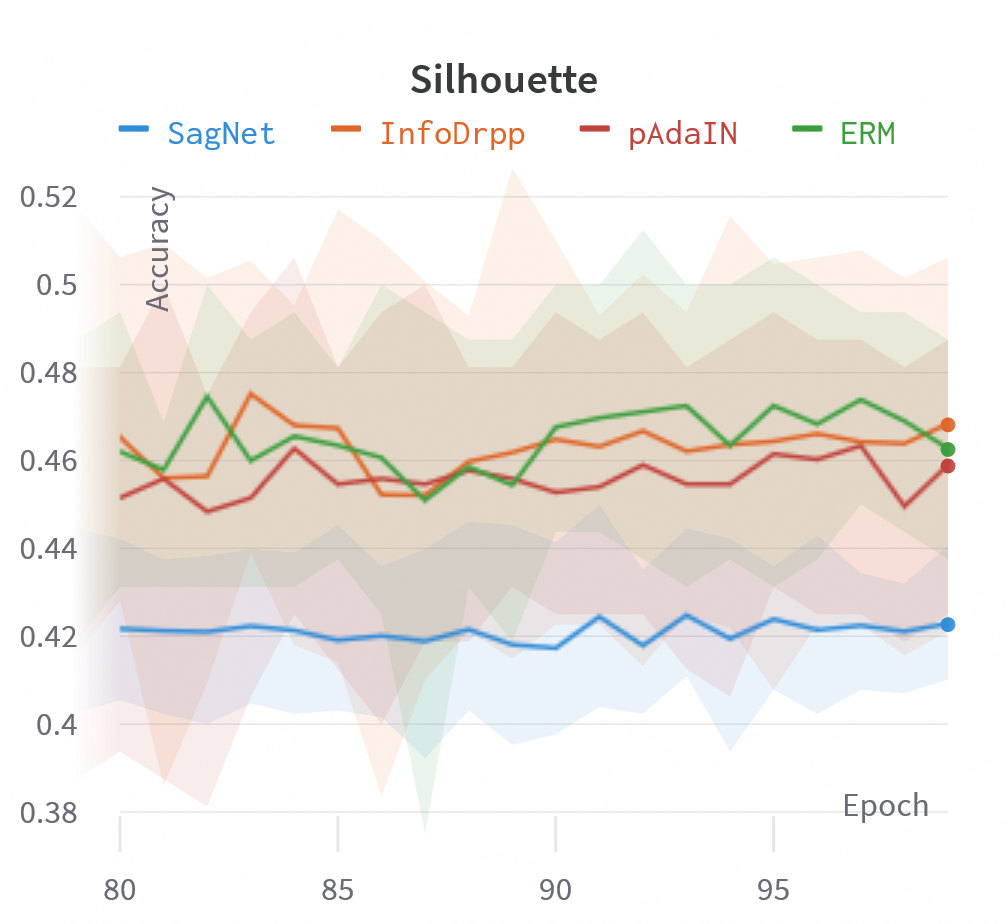}
     \end{subfigure}
        \caption{Accuracies of four different methods on three datasets. For each method, ten independent instances are trained and evaluated after every epoch. Lines denote average performance across then ten runs, shaded areas denote their standard deviation, highlighting significant performance fluctuations that should be taken into account.}
        \label{fig:testcurves}
\end{figure*}

\subsection{Current practice}

Existing literature about biased learning, and also about domain generalization in a broader sense, shares common evaluation practices. Algorithms are scored on different \ac{ood} datasets in terms of an evaluation metric, usually classification accuracy, and simply averaged across datasets. From a statistical perspective, this seemingly obvious practice has several deficiencies. 

First, and most importantly, often only a single point estimate of the metric is reported. However, deep learning algorithms are trained in highly stochastic fashion. Performance varies between independently trained models, since they almost certainly correspond to different local minima of the loss function. This effect is often particularly strong in the presence of domain shifts (\cf \cref{fig:testcurves}). It is evident that best practice would be to, at the very least, train multiple models with the same data and report their mean accuracy and its standard deviation. In the context of texture bias this becomes all the more important, as the variability between (and also within) training runs tends to be high.

Second, neural network training is an iterative process without a natural, unambiguous stopping criterion. Which iteration to regard as the "final" model to be evaluated is therefore invariably based on some model selection rule. Common strategies are to pick the one with the best in-domain validation score, to use a fixed iteration count, or to declare convergence based on some early-stopping rule \cite{zhang2005boosting, yao2007early, baker2017accelerating}. Looking at the texture and shape literature, we find a lack of information about the chosen strategy. This makes a fair comparison all but impossible: texture bias schemes tend to exhibit unusually high fluctuations between training epochs, such that a different model selection rule may reverse the relative performance of two methods. This concerns all tested methods and various datasets (see \cref{fig:testcurves} and \cref{tab:bestvslast}).
In this regard, we emphasize the efforts of \cite{gulrajani2020search}, who defined a set of strategies to account for performance fluctuations and encourage multiple training runs to collect the necessary statistics for fair comparisons over numerous datasets. A diverse spectrum of image domains helps to mitigate over-fitting towards particular domains and favours generic mechanisms over dataset-specific heuristics. To obtain statistically sound conclusions it is, however, not enough to simply average scores over different datasets, as done in \cite{gulrajani2020search}. Naturally, different datasets possess different characteristics and as such are potentially not commensurable, so that simple averaging becomes meaningless \cite{webb2000multiboosting}.
We instead propose a rigorous, statistically founded hypothesis testing framework to compensate for the differences.

\subsection{Hypothesis testing}
This section serves as background for our experiments. We emphasize that it summarizes well-established textbook knowledge, \eg \cite{lehmann1986testing}. 
The general aim is to compare $n$ algorithms $a_1, a_2, \dots, a_n$ on $m$ datasets $d_1, d_2, \dots, d_m$. We are interested in comparing the algorithms based on an evaluation metric $\mathcal{M}$, where $c_{ij}=\mathcal{M}(a_i, d_j)$ is the (averaged) score of algorithm $a_i$ on dataset $d_j$ (\eg, accuracy). Based on $c_{ij}$, we want to decide if the algorithms are statistically different \cite{demvsar2006statistical}. We assume that the underlying metric can be expressed additively by an effect due to the algorithm and an effect due to the dataset, and $c_{ij}$ is an estimate of this underlying metric. We define the null hypothesis and alternative hypothesis as following:
\begin{equation}
    \begin{split}
        H_0 : \forall ik\,\, \gamma(a_i) &= \gamma(a_k) \\
        H_1 : \exists ik\,\, \gamma(a_i) &\neq \gamma(a_k),
    \end{split}
\end{equation}
where $\gamma(a_i)$ is the effect of algorithm $a_i$. In case of $n=2$ and $m=1$, we have a two-sample setting  and with a little abuse of notation, we can write $d=\mathcal{M}(a_1, d) - \mathcal{M}(a_2, d)$ to test 
$H_0 : \gamma(a_1) = \gamma(a_2)$ versus $H_1 : \gamma(a_1) \neq \gamma(a_2)$.

The decision whether to reject the null hypothesis is based on the $p$-value, which is defined as the probability of obtaining a more extreme result than the one observed, assuming that $H_0$ is true:
\begin{equation}
    p \coloneqq P(D > d \mid H_0) + P( D < d \mid H_0),
\end{equation}
where $D$ is the test statistic associated to our observed difference $d$. Beforehand, we must choose a (user-defined) significance level $\alpha$ and reject $H_0$ if $p < \alpha$. The setup gives rise to two (inevitable) error types: \textit{Type I error} when rejecting $H_0$ although it is true, and \textit{Type II error} when not rejecting $H_0$ although it is false. Here, the Type I error probability is equal to $\alpha$. 

Generally, comparing multiple algorithms on multiple datasets follows a two-step procedure. First, an omnibus test is applied to the hypothesis of all algorithms performing equally well, \ie $H_0$ cannot be rejected. If $H_0$ can be rejected, a follow-up post-hoc test inspects each pair of algorithms to pinpoint where differences exist. A recommended omnibus test \cite{demvsar2006statistical} in the context of ML is the Friedman test \cite{friedman1937use}. It ranks all algorithms separately in each domain (\eg dataset), and uses the \emph{average rank} $R_i$ to calculate the Friedman statistic:
\begin{equation}
    \chi_F^2 = \frac{12m}{n(n+1)}\bigg[\sum_{i=1}^n R_i^2 - \frac{n(n+1)^2}{4}\bigg],
\end{equation}
which follows approximately a $\chi^2$ distribution with $(n-1)$ degrees of freedom. The Iman-Davenport extension to the Friedman statistic compensates for too conservative decisions and is defined as 
\begin{equation}
    F_F = \frac{(m-1)\chi_F^2}{m(n-1)-\chi_F^2},
\end{equation}
which is approximately distributed according to an $F$-distribution with $n-1$ and $(m-1)(n-1)$ degrees of freedom.
If the $p$-value is below the pre-defined significance threshold $\alpha$, the Nemenyi post-hoc test \cite{nemenyi1963distribution} compares pairs of differences.
The Nemenyi test statistic for algorithms $a_i$ and $a_j$ is defined as 
\begin{equation}
    z = |R_{i} - R_{j}| / \sqrt{\frac{m(m+1)}{6n}}.
\end{equation}
For large values of $z$ the difference is significant. 
Note that the two-step procedure is necessary to maintain the overall Type I error probability, which is not the case when testing all pairs within a two-sample framework at level $\alpha$. 

\section{BiasBed framework}\label{sec:BiasBedframework}

We introduce \name{}, a highly flexible and generic test suite to implement, train and evaluate existing algorithms in a rigorous manner. We currently include seven texture debiasing algorithms, ten datasets, different model selection strategies and an omnibus hypothesis test and a post-hoc hypothesis test. Adding existing and new algorithms, other datasets or running a hyperparameter search for one algorithm is a matter of a few lines. In fact, our framework provides a full integration with Weights and Biases \cite{wandb} providing the full range of logging, visualization and reports.

\paragraph{Algorithms.}
\name{} currently supports several algorithms that are tailored towards shape biased learning. Our baseline algorithm is a plain ResNet-50 trained with standard empirical risk minimization \cite{vapnik1999overview}. The same algorithm can be trained on Stylized ImageNet to duplicate the approach by Geirhos \etal \cite{geirhos2018imagenet}. Additional algorithms are: Shape-Texture Debiased Neural Network (Debiased, \cite{li2020shape}), Style-Agnostic Network (SagNet, \cite{nam2021reducing}), Permuted Adaptive Instance Normalization (pAdaIN, \cite{nuriel2021permuted}), Informative Dropout (InfoDrop, \cite{shi2020informative}) and Deep Augmentation (DeepAug, \cite{hendrycks2021many}). We show in the Appendix  how to easily add more algorithms besides the ones listed here.

\paragraph{Datasets.}

Besides the standard dataloader for ImageNet1k \cite{deng2009imagenet}, \name{} contains dataloaders for five additional texture-bias datasets and four robustness datasets. Stylized ImageNet \cite{geirhos2018imagenet} and Cue-Conflict \cite{geirhos2021partial} are stylized versions of ImageNet (1000 classes) and 16-class ImageNet, respectively, where the latter particularly uses texture cues for stylization. Datasets that contain pure shape information are Edges, Sketches and Silhouettes from \cite{geirhos2021partial}. We further extend BiasBed by including the following robustness datasets: ImageNet-A \cite{hendrycks2021nae}, ImageNet-R \cite{hendrycks2021many} and two versions of DeepAug~\cite{hendrycks2021many}.

\paragraph{Model selection.}

In \Cref{sec:evaluationpractices}, we have seen that when testing on out-of-distribution datasets, where performance fluctuations during training are rather high, the criterion used to pick the final model can have severe impacts on the final quality metrics. Therefore, we have implemented three model selection methods in \name{} to unify evaluation.

\begin{itemize}
    \item \textit{Best-epoch} is an oracle method that chooses the best test set score over all epochs after training for a fixed number of epochs. Importantly, this method is \textit{cherry-picking} and should be discouraged to use. We include it here to highlight the severe effects of model selection on evaluation.  
    \item \textit{Last-n-epochs} averages the test set scores over the last $n$ epochs, to mitigate the fluctuations observed in the \ac{ood} regime. \Ie, once the training has converged the model is evaluated after every training epoch, and a representative average score is reported.
    \item \textit{Best-training-validation} chooses a single test set score according that achieves the best \emph{in-domain} performance on the validation set. This method is the most rigorous one, but leads to the highest variance between independent test runs. To decrease that variance, more runs are needed. 
\end{itemize}

\paragraph{Evaluation.}

\name{} includes a full, rigorous evaluation protocol that can be run with a single command. It collects the result according to the defined model selection method, algorithms and datasets, and computes the averaged scores. In a second step, the Friedman test with Iman-Davenport extension \cite{iman1980approximations} is applied with a possible post-hoc Nemenyi test to identify significant differences. All results are reported back in a Pandas dataframe \cite{reback2020pandas, mckinney-proc-scipy-2010} or optionally in \LaTeX~tables (as those in \Cref{sec:experiments}).

\section{Experiments}\label{sec:experiments}

We run experiments for all implemented algorithms and extensively compare different model selection methods, highlighting the need for a common, explicit protocol. We use hypothesis testing to properly compare across algorithms. The following results are all generated by running a single command for each algorithm in \name{}. 

\subsection{Model selection}

\begin{table}[ht]
    \centering
    \footnotesize
    \begin{tabular}{@{}lccc@{}}
        \toprule
        Algorithm & Silhouette & Edge & Cue-Conflict \\
        \midrule
        ERM & 50.8 / 47.3 (\textcolor{red}{-3.5}) & 32.6 / 22.6 (\textcolor{red}{-10.0}) & 26.4 / 22.2 (\textcolor{red}{-4.2}) \\
        SagNet & 48.1 / 43.3 (\textcolor{red}{-4.8}) & 32.7 / 25.3 (\textcolor{red}{-7.4}) & 21.3 / 20.0 (\textcolor{red}{-1.3})\\
        pAdaIN & 48.0 / 43.8 (\textcolor{red}{-4.2}) & 29.0 / 22.3 (\textcolor{red}{-6.7}) & 24.0 / 21.5 (\textcolor{red}{-2.5})\\
        InfoDrop & 51.2 / 47.8 (\textcolor{red}{-3.4}) & 31.7 / 19.0 (\textcolor{red}{-11.7}) & 26.6 / 22.8 (\textcolor{red}{-3.8})\\
        \bottomrule
  \end{tabular}
    \caption{Comparison of best epoch (oracle) \vs checkpoint with best validation accuracy for selected algorithms and datasets.}
    \label{tab:bestvslast}
    \vspace{-1em}
\end{table}

We quantitatively elaborate on the findings of \cref{fig:testcurves}. We train each algorithm ten times and test on \textit{Silhouette}, \textit{Edge} and \textit{Cue-Conflict} after \textit{every} epoch during training. In \Cref{tab:bestval} we report the best accuracy over all epochs averaged across runs versus the last score received, averaged across all runs. Clearly, all algorithms drop severely in performance when one does not use an oracle model selection method, highlighting the need to properly define criteria how to select the model to be used. We emphasize that this step should always be part of an evaluation.

\subsection{Results}

\paragraph{Evaluation.}
We report results choosing the best validation performance for all algorithms and datasets in \Cref{tab:bestval} and choosing the last 30 epochs in \Cref{tab:lastn}. All results are once more gathered across multiple independent runs to account for randomness in training. In particular, we do not include an average column across dataset scores per algorithm, as discussed in \Cref{sec:evaluationpractices}. Note that in three cases data used for training was augmented in a similar way as the dataset used for testing, \ie there is no domain shift between train and test distribution. Naturally, performance is significantly higher. These results are \underline{underlined} to emphasize this fact.


\begin{table*}
 \centering
 \caption{Results choosing the best validation accuracy per run. Columns are grouped according to texture bias and adversarial robustness. Datasets marked with $\dagger$ are those where an algorithm has been trained on the corresponding dataset, \ie no distribution gap in the test set. The specific score is \underline{underlined}. Input augmented and feature augmented methods are grouped together.}
 \label{tab:bestval}
 \resizebox{\textwidth}{!}{
 \begin{tabular}{@{}l|c|ccccc|cccc@{}}
   \toprule
    Algorithm & ImageNet1k & Silhouette & Edge & Sketch & Cue-Conflict & Stylized ImageNet$^\dagger$ & ImageNet-A & ImageNet-R & DeepAug (CAE)$^\dagger$ & DeepAug (EDSR)$^\dagger$ \\
   \midrule
   ERM \cite{vapnik1999overview} & 73.8 $\pm$ 0.2 & 47.3 $\pm$ 2.4 & 22.6 $\pm$ 3.3 & 56.0 $\pm$ 1.0 & 22.2 $\pm$ 0.9 & 7.9 $\pm$ 0.2 & 2.0 $\pm$ 0.2 & 22.8 $\pm$ 0.4 & 44.4 $\pm$ 0.3 & 48.3 $\pm$ 0.5 \\ \hline
   pAdaIN \cite{nuriel2021permuted} & 73.2 $\pm$ 0.1 & 43.8 $\pm$ 2.2 & 22.3 $\pm$ 1.7 & 56.6 $\pm$ 0.8 & 21.5 $\pm$ 0.6 & 8.1 $\pm$ 0.1 & 1.4 $\pm$ 0.1 & 21.4 $\pm$ 0.3 & 42.8 $\pm$ 0.2 & 48.8 $\pm$ 0.3 \\
SagNet \cite{nam2021reducing} & 74.2 $\pm$ 0.5 & 43.3 $\pm$ 1.6 & 25.3 $\pm$ 2.0 & 59.2 $\pm$ 1.1 & 20.0 $\pm$ 0.2 & 6.2 $\pm$ 0.1 & 1.5 $\pm$ 0.2 & 21.8 $\pm$ 0.5 & 43.8 $\pm$ 0.4 & 47.7 $\pm$ 0.5 \\
InfoDrop \cite{shi2020informative} & 73.3 $\pm$ 0.2 & 47.8 $\pm$ 2.8 & 19.0 $\pm$ 4.3 & 56.7 $\pm$ 2.0 & 22.8 $\pm$ 0.6 & 7.9 $\pm$ 0.3 & 2.2 $\pm$ 0.1 & 22.7 $\pm$ 0.3 & 44.4 $\pm$ 0.2 & 48.6 $\pm$ 0.5 \\ \hline

Stylized ERM \cite{geirhos2018imagenet} & 55.9 $\pm$ 0.5 & 46.9 $\pm$ 2.8 & 58.4 $\pm$ 3.1 & 70.2 $\pm$ 1.4 & 53.7 $\pm$ 1.4 & \underline{53.2 $\pm$ 0.2} & 0.8 $\pm$ 0.1 & 25.0 $\pm$ 0.3 & 39.1 $\pm$ 0.5 & 40.4 $\pm$ 0.4 \\
Debiased \cite{li2020shape} & 74.4 $\pm$ 0.1 & 48.7 $\pm$ 2.8 & 30.8 $\pm$ 5.1 & 60.5 $\pm$ 1.2 & 28.9 $\pm$ 1.1 & 16.1 $\pm$ 0.3 & 2.7 $\pm$ 0.2 & 27.4 $\pm$ 0.4 & 49.6 $\pm$ 0.2 & 51.3 $\pm$ 0.3 \\
DAug. ERM (CAE) \cite{hendrycks2021many} & 73.7 $\pm$ 0.2 & 51.3 $\pm$ 3.2 & 34.7 $\pm$ 7.4 & 63.5 $\pm$ 2.6 & 29.9 $\pm$ 2.3 & 12.7 $\pm$ 2.0 & 2.6 $\pm$ 0.2 & 27.8 $\pm$ 1.7 & \underline{61.4 $\pm$ 5.8} & 55.8 $\pm$ 2.9 \\
DAug. ERM (EDSR) \cite{hendrycks2021many} & 72.8 $\pm$ 0.2 & 51.6 $\pm$ 1.3 & 31.7 $\pm$ 4.6 & 61.1 $\pm$ 1.5 & 32.4 $\pm$ 1.5 & 11.0 $\pm$ 0.3 & 2.0 $\pm$ 0.2 & 26.5 $\pm$ 0.5 & 52.3 $\pm$ 0.3 & \underline{65.1 $\pm$ 0.2} \\
   \bottomrule
  \end{tabular}
  }
\end{table*}


\begin{table*}
 \centering
 \caption{Results choosing the last 30 epochs per run. Columns are grouped according to texture bias and adversarial robustness. Datasets marked with $\dagger$ are those where an algorithm has been trained on the corresponding dataset, \ie no distribution gap in the test set. The specific score is \underline{underlined}. Input augmented and feature augmented methods are grouped together.}
 \label{tab:lastn}
 \resizebox{\textwidth}{!}{
 \begin{tabular}{@{}l|c|ccccc|cccc@{}}
   \toprule
    Algorithm & ImageNet1k & Silhouette & Edge & Sketch & CueConflict & Stylized ImageNet$^\dagger$ & ImageNet-A & ImageNet-R & DeepAug (CAE)$^\dagger$ & DeepAug (EDSR)$^\dagger$ \\
   \midrule
   ERM \cite{vapnik1999overview} & 73.3 $\pm$ 0.4 & 46.8 $\pm$ 2.4 & 22.6 $\pm$ 3.7 & 56.3 $\pm$ 1.3 & 22.1 $\pm$ 0.9 & 7.7 $\pm$ 0.3 & 2.1 $\pm$ 0.2 & 22.5 $\pm$ 0.5 & 43.8 $\pm$ 0.5 & 47.8 $\pm$ 0.6 \\ \hline
    pAdaIN \cite{nuriel2021permuted} & 73.1 $\pm$ 0.1 & 44.5 $\pm$ 2.1 & 22.1 $\pm$ 2.9 & 56.5 $\pm$ 0.9 & 21.4 $\pm$ 0.7 & 8.1 $\pm$ 0.2 & 1.5 $\pm$ 0.1 & 21.4 $\pm$ 0.3 & 42.7 $\pm$ 0.3 & 48.6 $\pm$ 0.3 \\
SagNet \cite{nam2021reducing} & 73.9 $\pm$ 0.5 & 44.6 $\pm$ 1.4 & 27.1 $\pm$ 2.4 & 60.4 $\pm$ 1.2 & 19.7 $\pm$ 0.5 & 6.2 $\pm$ 0.2 & 1.7 $\pm$ 0.2 & 22.5 $\pm$ 0.5 & 43.2 $\pm$ 0.7 & 47.3 $\pm$ 0.5 \\
InfoDrop \cite{shi2020informative} & 72.9 $\pm$ 0.5 & 47.0 $\pm$ 2.7 & 18.9 $\pm$ 4.4 & 56.6 $\pm$ 1.7 & 22.9 $\pm$ 0.8 & 7.7 $\pm$ 0.3 & 2.2 $\pm$ 0.2 & 22.7 $\pm$ 0.5 & 43.9 $\pm$ 0.6 & 48.4 $\pm$ 0.6 \\ \hline
Stylized ERM \cite{geirhos2018imagenet} & 55.3 $\pm$ 0.7 & 47.3 $\pm$ 2.6 & 59.5 $\pm$ 3.5 & 70.2 $\pm$ 1.2 & 54.2 $\pm$ 1.5 & \underline{52.4 $\pm$ 0.6} & 0.7 $\pm$ 0.1 & 25.0 $\pm$ 0.5 & 38.8 $\pm$ 0.7 & 40.1 $\pm$ 0.6 \\
Debiased \cite{li2020shape} & 74.0 $\pm$ 0.3 & 48.3 $\pm$ 2.8 & 29.4 $\pm$ 4.9 & 60.1 $\pm$ 1.3 & 29.1 $\pm$ 1.1 & 15.6 $\pm$ 0.5 & 2.6 $\pm$ 0.2 & 27.2 $\pm$ 0.5 & 49.1 $\pm$ 0.5 & 50.8 $\pm$ 0.6 \\
DAug. ERM (CAE) \cite{hendrycks2021many} & 72.8 $\pm$ 0.3 & 50.5 $\pm$ 3.4 & 33.1 $\pm$ 7.3 & 62.6 $\pm$ 2.9 & 29.4 $\pm$ 2.4 & 12.4 $\pm$ 1.9 & 2.7 $\pm$ 0.2 & 27.0 $\pm$ 1.7 & \underline{60.6 $\pm$ 5.9} & 55.2 $\pm$ 2.9 \\
DAug. ERM (EDSR) \cite{hendrycks2021many} & 71.9 $\pm$ 0.7 & 51.4 $\pm$ 2.2 & 32.0 $\pm$ 4.2 & 60.4 $\pm$ 1.9 & 32.3 $\pm$ 1.2 & 10.6 $\pm$ 0.5 & 2.0 $\pm$ 0.2 & 25.8 $\pm$ 0.6 & 51.1 $\pm$ 0.9 & \underline{64.2 $\pm$ 0.8} \\
   \bottomrule
  \end{tabular}
  }
\end{table*}

\paragraph{Hypothesis testing.}

We conduct the Friedman test on the scores of \Cref{tab:bestval}. We conduct hypothesis tests in two ways: once with all datasets, and another more strict setting where we reject datasets which were created using the same methods as for training any of the algorithms. We chose a significance level of $\alpha = 0.05$ and test for significance. The returned F-statistic is $7.46$ with an uncorrected $p$-value of $0.000002$ when using all datasets and $F=4.98$, $p=0.00045$ in the strict setting. In both cases, $p < \alpha$ and therefore we reject $H_0$. In \Cref{tab:posthoc_full} and \Cref{tab:posthoc_strict} we report all pairwise comparisons of algorithms using the Nemenyi post-hoc test for both settings.


\begin{table*}
 \centering
 \footnotesize
 \caption{Full post-hoc Nemenyi test based on validation. For this hypothesis test we use scores from all datasets.}
 \label{tab:posthoc_full}
 \begin{tabular}{@{}lcccccccc@{}}
   \toprule
    Algorithm & ERM & pAdaIN & SagNet & InfoDrop & Stylized ERM & Debiased & DAug. ERM (CAE) & DAug. ERM (EDSR) \\
   \midrule
   ERM \cite{vapnik1999overview} & 1.0 & 0.9 & 0.9 & 0.9 & 0.9 & 0.211 & 0.053 & 0.478 \\
pAdaIN \cite{nuriel2021permuted} & & 1.0 & 0.9 & 0.9 & 0.642 & 0.012 & 0.002 & 0.053 \\
SagNet  \cite{nam2021reducing} & & & 1.0 & 0.9 & 0.806 & 0.03 & 0.005 & 0.111 \\
InfoDrop \cite{shi2020informative} & & & & 1.0 & 0.9 & 0.254 & 0.069 & 0.533 \\
Stylized ERM \cite{geirhos2018imagenet} & & & & & 1.0 & 0.642 & 0.303 & 0.9 \\
Debiased \cite{li2020shape} & & & & & & 1.0 & 0.9 & 0.9 \\
DAug. ERM (CAE) \cite{hendrycks2021many} & & & & & & & 1.0 & 0.9 \\
DAug. ERM (EDSR) \cite{hendrycks2021many} & & & & & & & & 1.0 \\
   \bottomrule
  \end{tabular}
\end{table*}


\begin{table*}
 \centering
 \footnotesize
 \caption{Partial post-hoc Nemenyi test based on validation. This hypothesis test is based solely on datasets that have a true distribution shift wrt. the training data.}
 \label{tab:posthoc_strict}
 \begin{tabular}{@{}lcccccccc@{}}
   \toprule
    Algorithm & ERM & pAdaIN & SagNet & InfoDrop & Stylized ERM & Debiased & DAug. ERM (CAE) & DAug. ERM (EDSR) \\
   \midrule
   ERM \cite{vapnik1999overview} & 1.0 & 0.897 & 0.9 & 0.9 & 0.9 & 0.505 & 0.241 & 0.897 \\
pAdaIN \cite{nuriel2021permuted} &  & 1.0 & 0.9 & 0.897 & 0.364 & 0.024 & 0.005 & 0.149 \\
SagNet  \cite{nam2021reducing} &  &  & 1.0 & 0.9 & 0.897 & 0.241 & 0.086 & 0.636 \\
InfoDrop \cite{shi2020informative} &  &  &  & 1.0 & 0.9 & 0.505 & 0.241 & 0.897 \\
Stylized ERM \cite{geirhos2018imagenet} &  &  &  &  & 1.0 & 0.9 & 0.766 & 0.9 \\
Debiased \cite{li2020shape} &  &  & & & & 1.0 & 0.9 & 0.9 \\
DAug. ERM (CAE) \cite{hendrycks2021many} &  &  & & & & & 1.0 & 0.9 \\
DAug. ERM (EDSR) \cite{hendrycks2021many} &  &  & & & & & & 1.0 \\
   \bottomrule
  \end{tabular}
  \vspace{-0.5em}
\end{table*}

\section{Discussion}\label{sec:discussion}

Our results, found in Tabs. \ref{tab:bestvslast}, \ref{tab:bestval}, \ref{tab:lastn}, and most importantly \cref{tab:posthoc_full} and \cref{tab:posthoc_strict}, allow us to draw several conclusions that support the usage of the formal hypothesis testing framework for this evaluation.

\paragraph{Input augmentation and feature augmentation.} The considered algorithms can be grouped into two main families: those that focus on augmenting the inputs that go into the network, and those that focus on the activations inside the network. We observe in \cref{tab:bestval} and \cref{tab:lastn} that algorithms that perform input augmentation (Stylized ERM, Debiased and both DeepAug ERMs) tend to have larger differences in performance \wrt ERM (trained solely on ImageNet) than algorithms that try to mitigate style bias through latent space (feature) augmentation (SagNet, InfoDrop, pAdaIN). This suggests that the proposed feature augmentations -- such as using AdaIN to shift and scale feature maps -- fail to capture real variations in texture. We hypothesize that such augmentations require the decoder from a pre-trained auto-encoder to better express texture features. However, and importantly, only in rare cases do the algorithms statistically differ from each other (\cf \cref{tab:posthoc_full}, \cref{tab:posthoc_strict}). In fact, taking into account only the true OOD datasets none of the implemented algorithms significantly outperforms a baseline ERM.

\vspace{-1em}
\paragraph{Model selection criteria.} If model selection is done based on validation accuracy, the margins are extremely small compared to the uncertainty, as measured by the standard deviation over 10 different runs of an algorithm. This suggests that, while validation performance is of course to some degree predictive of OOD performance, it apparently falls short of capturing all relevant effects that impact such generalization, and is not sufficient. It seems clear that having an explicit, formal model selection strategy is of paramount importance. We can also conclude that reporting results for a single run is not a reliable way of comparing different approaches, as the stochastic nature of the training alone can flip the relative performance of two methods. 

\vspace{-1em}
\paragraph{Intra-run and inter-run variability.} While some authors acknowledge variability in the results across different runs, it is rare to find statements about intra-run variability, \ie, strong performance fluctuations between nearby training checkpoints, as is depicted by the shaded regions in \cref{fig:testcurves}. It is sometimes the case that authors report the mean and standard deviation of performance metrics for several runs to mitigate this, but the significant variations not only between independent training runs, but also among different epochs of an apparently converged training mean that averages without the associated uncertainty are problematic, and it is nearly impossible to draw reliable conclusions without a formal framework. The complex interplay between these factors needs to be acknowledged and analysed more closely. These types of variability stem from different sources and need to be handled in different ways.

\vspace{-1em}
\paragraph{Importance of hypothesis testing.} The presented results also highlight the importance of using a formal comparison framework when dealing with such complex cases. When informally analysing Tables \ref{tab:bestval} and \ref{tab:lastn}, such as by simply averaging each algorithm's performance across datasets, it is easy to arrive at erroneous conclusions based on spurious results. For instance, the results seen in \cref{tab:bestval} could lead one to believe that the Debiased and DeepAug algorithms do outperform competitors due to their high performance on datasets such as Sketch, CueConflict, and Sylized IN, with little or no performance loss on ImageNet1k when compared to ERM, the baseline result. But this analysis does not take into account the variances of these results and the complex interplay between the different measured accuracies, sample sizes, \etc. In fact, the results of the post-hoc Nemenyi test, reported in \cref{tab:posthoc_full} and \cref{tab:posthoc_strict}, tell us that these algorithms do not differ from ERM in a statistically significant way -- one can not refute that null hypothesis that the two methods are on par. It is possible that using more datasets would allow us to identify which algorithms, if any, have a statistically significant effect, but even with the rather many runs in our test bed, no staistically supported difference has been observed yet, and we cannot confidently establish a correct ranking.


\vspace{-1em}
\paragraph{Interpretation of hypothesis testing.} When using hypothesis test, one needs to be precise \wrt the conclusions drawn from the results. In particular, rejecting the null hypothesis means that if the null hypothesis were correct, we would almost certainly (with significance $\alpha$) not observe a so extreme (or even more extreme) difference. However, even if the significance level is high we cannot conclude superiority of one algorithm over another for an unseen domain. The \textit{significant} difference is only valid under the experimental setting, \ie within the datasets and algorithms used. In turn, being unable to rejecting the null hypothesis should not be misinterpreted as a proof of equality of algorithms. 

\section{Conclusion}\label{sec:conclusion}

When analyzing existing methods tailored towards texture-free learning, common datasets, evaluation protocols and reports are missing. In this work, we introduced \name{} to alleviate the aforementioned limitations. In particular, we have seen that model selection methods play a critical role in OOD testing and fair comparisons are only possible if algorithms are evaluated in a rigorous fashion. Hypothesis testing can fill this gap by providing statistically sound comparisons. Moreover, one must be careful to draw the right conclusions, \eg, low significance of a difference does not necessarily mean that two methods perform on par, but may also indicate that there are too few observations to make a confident statement about their difference. Our framework provides the necessary tools to implement, test and compare existing and new algorithms. Our intention is not to make negative claims or invalidate any particular approach. Rather, we hope to encourage the community to leverage existing statistical expertise and ensure fair and rigorous quantitative evaluations that drive forward the field.

\section{Broader impact}\label{sec:broaderimpact}


The aim of the work presented in this paper is to provide a solid foundation for future research on style bias of neural networks. Such hypothesis testing framework are commonplace in other fields where the effects of different experiments can only be observed in noisy results, such as many areas of physics, medicine and psychology. We hope that the presented framework will be used in the future by other researchers through the openly released codebase to find and validate novel algorithms that mitigate texture bias. Furthermore, the proposed methodology -- and codebase -- can be used to perform rigorous testing of algorithms for solving other computer vision and machine learning problems, such as domain adaptation and domain generalization which are notably hard for the validation of different algorithms. It is our belief that setting a higher bar and expecting rigorous testing from authors who propose novel methods will, in the long run, improve the quality of research in this field.

{\small
\bibliographystyle{ieee_fullname}
\bibliography{egbib}
}

\include{supplementary}

\end{document}

%% file: supplementary.tex
\newpage

\onecolumn
\appendix
\appendixpage
\addappheadtotoc

\section{BiasBed}
\name{} is a Python package that can simply be installed with pip package manager. Once installed, we can run sets of experiments with a single command. We follow by default a common training protocol: the learning rate is set to $1e\text{-}4$, the learning decay after 30, 60 and 90 epochs is $0.1$, the SGD optimizer has momentum $0.9$ with weight decay $1e\text{-}4$, and we train for 100 epochs in total. We use ResNet-50 as a backbone model for all experiments.

\subsection{Adding algorithms}
New algorithms in \name{} can easily be added by extending the provided abstract algorithm class. The important function to implement here is \verb|update(x, y)|. This update function receives batches of input data and corresponding ground truth. It needs to predict the logits, compute the loss and backpropagate the gradients. To implement empirical risk minimization \cite{vapnik1999overview}, for example, we add a folder \verb|ERM| in the algorithms folder and add \verb|algorithm.py| with the corresponding \verb|update(x, y)| function:

\usemintedstyle{fruity}
\begin{minted}[frame=single, bgcolor=bg]{python}
def update(self, x, y):
    self.optimizer.zero_grad(set_to_none=True)
    with autocast(enabled=self.algorithm_cfg.mixedprec):
        logits = self.model(x)
        loss = self.loss(logits, y)

    # Backward loss
    self.scaler.scale(loss).backward()
    self.scaler.step(self.optimizer)
    self.scaler.update()
\end{minted}

Each algorithm folder contains two additional config files \verb|config.yaml| and \verb|sweep.yaml|. The former includes all algorithm specific (hyper-) parameters:
\usemintedstyle{fruity}
\begin{minted}[frame=single, bgcolor=bg]{yaml}
# Include default parameters of your algorithm here.
mixedprec: True
backbone: resnet50 # Net for MNIST
optimizer: SGD
learning_rate: 1e-4
milestones:
  - 30
  - 60
  - 90
momentum: 0.9
weight_decay: 1e-4
gamma: 0.1
\end{minted}
and the latter includes all (hyper-) parameters necessary for sweeping over individual parameters in the algorithm and main config files, \eg
\begin{minted}[frame=single, bgcolor=bg]{yaml}
# Include sweep parameters of your algorithm here
parameters:
  epoch:
    values:
      - 10
      - 20
      - 30
  momentum:
    values:
      - 0.7
      - 0.8
      - 0.9
\end{minted}
\name{} takes care of registering and adding the algorithm to the framework. See \Cref{sec:runningbiasbed} on how to run experiments with the newly added algorithm.

\subsection{Adding datasets}
Adding dataloaders for datasets is equally simple. We need to add a folder with the dataloader name and implement the \verb|train_loader, val_loader, test_loader| from the dataloader template in a file called \verb|dataloader.py|, \eg in the case of Cue-Conflict
\usemintedstyle{fruity}
\begin{minted}[frame=single, bgcolor=bg]{python}
def train_loader(self) -> Iterable:
    data_loader = DataLoader(self.dataset,
                             batch_size=self.config.training.batch_size,
                             num_workers=self.config.num_workers,
                             sampler=sampler)
    return data_loader
\end{minted}
where \verb|self.dataset| is a PyTorch \verb|ImageFolder| dataset. Of course, each dataset comes with its own config file to include dataset specific (hyper-) parameters. Once the dataloader is added, the dataset can be seamlessly added to the main config file detailed in the following section.

\subsection{Running BiasBed}\label{sec:runningbiasbed}

\name{} supports various training modes, including full support of half precision, multi-GPU or hyperparameter sweeping with cluster support. We provide a fully flexible main configuration file to activate or deactivate settings. Only a single line has to be edited for training an algorithm on a single GPU or on a compute node with multiple GPUs. We fully integrated Weights and Biases \cite{wandb} into our framework, too, such that these are automatically used to search and tune hyperparameters. A \name{} user can either use our code to launch runs on common high performance computing environments or easily add a custom launcher.  
We can start a single run with the command:
\usemintedstyle{fruity}
\begin{minted}[frame=single, bgcolor=bg]{python}
biasedbed single # Starts a single run with the algorithm set in config.yaml
\end{minted}
or start automated hyperparameter tuning with 
\usemintedstyle{fruity}
\begin{minted}[frame=single, bgcolor=bg]{python}
biasedbed sweep # Starts a sweep with settings from sweeping/sweep.yaml 
\end{minted}
In both cases, algorithm and dataset settings can be edited and activated in \verb|config.yaml|. For example, we can train SagNet \cite{nam2021reducing} with eight GPUs on standard ImageNet and evaluate the model on six different datasets with the following configuration:
\begin{minted}[frame=single, bgcolor=bg]{yaml}
# Distributed training
num_workers: 8
distributed: 1
world_size: 8
# Algorithm
algorithm:
  name: SagNet

# Training
training:
  dataset:
    name: ImageNet1k
  epochs: 100

# Testing
testing:
  datasets:
    - CueConflict
    - Silhouette
    - Sketch
    - Edge
    - ImageNetStylized
    - ImageNet1k
  interval: 1

\end{minted}

\section{Algorithms}

\paragraph{ERM} ``Empirical Risk Minimization" \cite{vapnik1999overview} minimizes the cross entropy loss across the training data and serves as our baseline algorithm. 

\paragraph{Stylized ImageNet} ``Imagenet-trained CNNs are biased towards texture; increasing shape bias improves accuracy and robustness" \cite{geirhos2018imagenet} is the first paper to recognize and rigorously demonstrate texture bias in existing neural architectures. To reduce texture bias, the authors propose a stylized version of ImageNet, where they use AdaIN \cite{huang2017arbitrary} to change the texture of one image with another random image of ImageNet. 

\paragraph{Debiased} ``Shape-texture debiased neural network training" \cite{li2020shape} extends the idea of \cite{geirhos2018imagenet} by augmenting the dataset online, \ie when feeding a batch of (original ImageNet) images into the network. Instead of only training on the content label, a convex combination of the style image class label and the content class label is used to guide the network to ``debiased" weights, \ie the network is forced to predict the content class solely from shape cues and the style class solely from texture cues. The authors argue, that performance is generally higher on all tested datasets (and not only on shape-biased sets) compared to \cite{geirhos2018imagenet}. 

\paragraph{DeepAugment} ``The Many Faces of Robustness: A Critical Analysis of Out-of-Distribution Generalization" \cite{hendrycks2021many} introduces additional deep augmentation techniques similar to \cite{geirhos2018imagenet}. In DeepAugment, an image is passed through an image-to-image network, but the forward pass is distorted by an altering the network. This distorts the resulting image in a similar way to augmentation methods. The authors defined a number of pertubations such as zeroing, negating, convolving, transposing, or switching activation functions and drew per image random samples from them. The networks used in DeepAugment are the pre-trained networks EDSR by \cite{EDSR} and CAE \cite{CAE}. The resulting augmented images for ImageNet are provided at \url{https://github.com/hendrycks/imagenet-r/tree/master/DeepAugment}. In principle, they can be combined with any other algorithm by appending them to the standard ImageNet dataset.

\paragraph{InfoDrop} ``Informative Dropout for Robust Representation Learning: A Shape-bias Perspective" \cite{shi2020informative} proposes an agnostic light-weight method to reduce texture bias in neural networks. The main idea is to enforce visual primitives such as edges and corners (\ie, regions with high shape information) and to reduce homogeneous and repetitive patterns (\ie, regions with low shape information). During training, neurons corresponding to input patches with low shape information are more likely to be zeroed out than neurons with high shape information patches. 

\paragraph{SagNet} ``Reducing Domain Gap by Reducing Style Bias" \cite{nam2021reducing} introduces a style-agnostic network that becomes invariant to texture with a style randomization and content randomization network. Features of a shared encoder are randomly interpolated with style features from another image in the batch. The style network is forced to predict the correct style label and the content network needs to predict the correct content label. The gradients of the former network are adversarially used to update the shared encoder. 

\paragraph{pAdaIN} ``Permuted AdaIN: Reducing the Bias Towards Global Statistics in Image Classification" \cite{nuriel2021permuted} follows a similar idea as SagNet but only incorporates a single style network that is forced to predict the correct content label from style-interpolated features.

\section{Full BiasBed results}
We report all individual results per algorithm according to the best in-domain validation score and average score over the last 30 epochs.

\subsection{Model selection: average of last 30 epochs}

\subsubsection{ERM \cite{vapnik1999overview}}
\begin{table}[H]
 \centering
 \resizebox{\textwidth}{!}{
 \begin{tabular}{@{}lcccccccccc@{}}
   \toprule
    Run & ImageNet1k & Silhouette & Edge & Sketch & CueConflict & ImageNetStylized & ImageNetA & ImageNetR & DeepAugCAE & DeepAugEDSR \\
   \midrule
   	 1 & 73.2 $\pm$ 0.3 & 47.2 $\pm$ 1.4 & 23.9 $\pm$ 1.7 & 56.4 $\pm$ 0.9 & 22.0 $\pm$ 0.6 & 7.6 $\pm$ 0.2 & 2.0 $\pm$ 0.1 & 21.8 $\pm$ 0.5 & 43.5 $\pm$ 0.6 & 47.6 $\pm$ 0.5 \\
	 2 & 73.1 $\pm$ 0.3 & 45.4 $\pm$ 2.4 & 28.9 $\pm$ 2.5 & 56.7 $\pm$ 0.8 & 22.8 $\pm$ 0.7 & 7.6 $\pm$ 0.3 & 1.7 $\pm$ 0.1 & 22.6 $\pm$ 0.4 & 43.7 $\pm$ 0.5 & 47.5 $\pm$ 0.5 \\
	 3 & 73.6 $\pm$ 0.3 & 45.6 $\pm$ 2.5 & 19.4 $\pm$ 1.8 & 55.4 $\pm$ 1.0 & 22.6 $\pm$ 0.7 & 7.5 $\pm$ 0.2 & 2.1 $\pm$ 0.2 & 22.8 $\pm$ 0.4 & 43.8 $\pm$ 0.5 & 48.0 $\pm$ 0.5 \\
	 4 & 73.2 $\pm$ 0.3 & 47.6 $\pm$ 2.0 & 17.9 $\pm$ 2.0 & 55.5 $\pm$ 1.1 & 21.6 $\pm$ 0.7 & 8.0 $\pm$ 0.3 & 2.2 $\pm$ 0.2 & 22.3 $\pm$ 0.4 & 43.8 $\pm$ 0.5 & 47.5 $\pm$ 0.4 \\
	 5 & 73.5 $\pm$ 0.3 & 46.3 $\pm$ 1.3 & 24.5 $\pm$ 2.8 & 56.4 $\pm$ 1.0 & 22.1 $\pm$ 0.8 & 7.7 $\pm$ 0.3 & 2.1 $\pm$ 0.1 & 22.6 $\pm$ 0.4 & 44.1 $\pm$ 0.6 & 48.7 $\pm$ 0.4 \\
	 6 & 73.2 $\pm$ 0.3 & 47.1 $\pm$ 2.2 & 21.0 $\pm$ 2.4 & 55.6 $\pm$ 1.3 & 21.3 $\pm$ 0.4 & 7.7 $\pm$ 0.3 & 2.2 $\pm$ 0.2 & 22.8 $\pm$ 0.3 & 43.8 $\pm$ 0.5 & 47.8 $\pm$ 0.4 \\
	 7 & 73.3 $\pm$ 0.3 & 46.1 $\pm$ 2.5 & 23.6 $\pm$ 2.3 & 57.5 $\pm$ 1.1 & 21.4 $\pm$ 0.8 & 7.8 $\pm$ 0.2 & 2.0 $\pm$ 0.2 & 22.5 $\pm$ 0.3 & 43.6 $\pm$ 0.6 & 48.0 $\pm$ 0.5 \\
	 8 & 73.3 $\pm$ 0.4 & 48.5 $\pm$ 1.8 & 21.0 $\pm$ 2.4 & 56.6 $\pm$ 1.2 & 23.1 $\pm$ 0.8 & 7.6 $\pm$ 0.2 & 2.1 $\pm$ 0.2 & 22.5 $\pm$ 0.5 & 44.0 $\pm$ 0.5 & 47.4 $\pm$ 0.6 \\
	 9 & 73.5 $\pm$ 0.3 & 48.9 $\pm$ 1.5 & 21.7 $\pm$ 2.1 & 56.2 $\pm$ 1.0 & 22.2 $\pm$ 0.6 & 7.5 $\pm$ 0.2 & 2.1 $\pm$ 0.1 & 22.3 $\pm$ 0.3 & 44.1 $\pm$ 0.5 & 47.5 $\pm$ 0.4 \\
	 10 & 73.5 $\pm$ 0.3 & 45.4 $\pm$ 2.6 & 23.7 $\pm$ 2.4 & 57.1 $\pm$ 1.3 & 22.4 $\pm$ 0.6 & 7.5 $\pm$ 0.2 & 2.0 $\pm$ 0.1 & 22.7 $\pm$ 0.4 & 44.0 $\pm$ 0.4 & 48.1 $\pm$ 0.4 \\
   \bottomrule
  \end{tabular}
  }
\end{table}

\subsubsection{Debiased \cite{li2020shape}}
\begin{table}[H]
 \centering
 \resizebox{\textwidth}{!}{
 \begin{tabular}{@{}lcccccccccc@{}}
   \toprule
    Run & ImageNet1k & Silhouette & Edge & Sketch & CueConflict & ImageNetStylized & ImageNetA & ImageNetR & DeepAugCAE & DeepAugEDSR \\
   \midrule
   	 1 & 74.0 $\pm$ 0.3 & 49.3 $\pm$ 3.0 & 32.7 $\pm$ 2.6 & 59.8 $\pm$ 0.8 & 29.4 $\pm$ 0.5 & 15.6 $\pm$ 0.4 & 2.6 $\pm$ 0.2 & 27.5 $\pm$ 0.4 & 49.3 $\pm$ 0.3 & 51.0 $\pm$ 0.5 \\
	 2 & 73.9 $\pm$ 0.3 & 48.7 $\pm$ 1.9 & 20.1 $\pm$ 1.3 & 61.1 $\pm$ 1.1 & 28.7 $\pm$ 0.8 & 15.4 $\pm$ 0.5 & 2.4 $\pm$ 0.2 & 27.0 $\pm$ 0.6 & 49.0 $\pm$ 0.6 & 51.0 $\pm$ 0.5 \\
	 3 & 74.1 $\pm$ 0.3 & 48.0 $\pm$ 1.8 & 36.3 $\pm$ 2.0 & 60.2 $\pm$ 1.0 & 29.6 $\pm$ 0.8 & 15.8 $\pm$ 0.5 & 2.5 $\pm$ 0.2 & 27.4 $\pm$ 0.5 & 49.0 $\pm$ 0.5 & 51.2 $\pm$ 0.4 \\
	 4 & 74.0 $\pm$ 0.3 & 46.1 $\pm$ 1.9 & 27.3 $\pm$ 2.3 & 60.8 $\pm$ 1.0 & 30.3 $\pm$ 0.5 & 15.9 $\pm$ 0.4 & 2.5 $\pm$ 0.2 & 27.0 $\pm$ 0.4 & 49.1 $\pm$ 0.5 & 50.5 $\pm$ 0.5 \\
	 5 & 73.7 $\pm$ 0.3 & 47.6 $\pm$ 1.6 & 32.7 $\pm$ 3.0 & 59.3 $\pm$ 1.3 & 29.1 $\pm$ 0.5 & 15.3 $\pm$ 0.4 & 2.5 $\pm$ 0.1 & 26.9 $\pm$ 0.4 & 48.9 $\pm$ 0.6 & 50.9 $\pm$ 0.5 \\
	 6 & 74.1 $\pm$ 0.4 & 46.9 $\pm$ 2.0 & 27.7 $\pm$ 2.6 & 60.6 $\pm$ 0.9 & 29.6 $\pm$ 0.7 & 15.8 $\pm$ 0.5 & 2.7 $\pm$ 0.2 & 27.7 $\pm$ 0.4 & 49.6 $\pm$ 0.5 & 51.1 $\pm$ 0.5 \\
	 7 & 74.0 $\pm$ 0.4 & 44.5 $\pm$ 2.3 & 31.2 $\pm$ 2.4 & 59.0 $\pm$ 1.1 & 27.1 $\pm$ 0.5 & 15.4 $\pm$ 0.5 & 2.5 $\pm$ 0.2 & 27.1 $\pm$ 0.3 & 49.0 $\pm$ 0.5 & 50.8 $\pm$ 0.5 \\
	 8 & 73.9 $\pm$ 0.3 & 50.5 $\pm$ 1.5 & 31.6 $\pm$ 3.3 & 60.0 $\pm$ 1.1 & 28.9 $\pm$ 0.8 & 15.9 $\pm$ 0.5 & 2.7 $\pm$ 0.1 & 27.0 $\pm$ 0.3 & 49.0 $\pm$ 0.5 & 51.0 $\pm$ 0.5 \\
	 9 & 73.9 $\pm$ 0.3 & 50.2 $\pm$ 2.0 & 27.1 $\pm$ 2.4 & 61.3 $\pm$ 0.9 & 30.2 $\pm$ 0.6 & 15.7 $\pm$ 0.4 & 2.5 $\pm$ 0.2 & 27.4 $\pm$ 0.5 & 49.2 $\pm$ 0.5 & 50.6 $\pm$ 0.6 \\
	 10 & 74.0 $\pm$ 0.3 & 51.2 $\pm$ 1.2 & 27.3 $\pm$ 1.9 & 59.4 $\pm$ 0.8 & 28.0 $\pm$ 0.5 & 15.4 $\pm$ 0.5 & 2.7 $\pm$ 0.2 & 27.1 $\pm$ 0.4 & 49.1 $\pm$ 0.6 & 50.3 $\pm$ 0.6 \\
   \bottomrule
  \end{tabular}
  }
\end{table}

\subsubsection{DeepAug ERM (CAE)\cite{hendrycks2021many}} 
\begin{table}[H]
 \centering
 \resizebox{\textwidth}{!}{
 \begin{tabular}{@{}lcccccccccc@{}}
   \toprule
    Run & ImageNet1k & Silhouette & Edge & Sketch & CueConflict & ImageNetStylized & ImageNetA & ImageNetR & DeepAugCAE & DeepAugEDSR \\
   \midrule
   	 1 & 72.9 $\pm$ 0.3 & 52.5 $\pm$ 1.3 & 32.8 $\pm$ 2.4 & 62.5 $\pm$ 1.1 & 30.9 $\pm$ 0.7 & 12.9 $\pm$ 0.3 & 2.8 $\pm$ 0.1 & 27.9 $\pm$ 0.5 & 62.6 $\pm$ 0.4 & 56.2 $\pm$ 0.5 \\
	 2 & 72.9 $\pm$ 0.3 & 51.1 $\pm$ 1.4 & 34.8 $\pm$ 1.8 & 64.0 $\pm$ 1.1 & 30.6 $\pm$ 0.6 & 13.0 $\pm$ 0.2 & 2.8 $\pm$ 0.1 & 27.7 $\pm$ 0.4 & 62.6 $\pm$ 0.3 & 55.8 $\pm$ 0.4 \\
	 3 & 72.5 $\pm$ 0.3 & 51.3 $\pm$ 1.2 & 39.2 $\pm$ 2.9 & 64.4 $\pm$ 1.3 & 29.4 $\pm$ 0.7 & 13.0 $\pm$ 0.3 & 2.6 $\pm$ 0.1 & 27.5 $\pm$ 0.4 & 62.3 $\pm$ 0.4 & 55.8 $\pm$ 0.4 \\
	 4 & 73.0 $\pm$ 0.2 & 50.5 $\pm$ 1.7 & 29.1 $\pm$ 2.0 & 65.0 $\pm$ 0.7 & 30.3 $\pm$ 0.7 & 12.9 $\pm$ 0.2 & 2.9 $\pm$ 0.1 & 27.6 $\pm$ 0.4 & 62.6 $\pm$ 0.3 & 56.0 $\pm$ 0.4 \\
	 5 & 72.8 $\pm$ 0.3 & 51.0 $\pm$ 1.7 & 32.1 $\pm$ 3.1 & 65.0 $\pm$ 0.9 & 29.3 $\pm$ 0.8 & 13.0 $\pm$ 0.2 & 2.6 $\pm$ 0.1 & 27.3 $\pm$ 0.4 & 62.4 $\pm$ 0.4 & 56.2 $\pm$ 0.4 \\
	 6 & 73.0 $\pm$ 0.3 & 53.4 $\pm$ 1.9 & 39.6 $\pm$ 1.4 & 63.3 $\pm$ 1.0 & 30.6 $\pm$ 0.8 & 12.8 $\pm$ 0.2 & 2.7 $\pm$ 0.1 & 28.1 $\pm$ 0.4 & 62.7 $\pm$ 0.4 & 56.5 $\pm$ 0.3 \\
	 7 & 72.7 $\pm$ 0.3 & 53.7 $\pm$ 1.6 & 36.6 $\pm$ 2.2 & 62.8 $\pm$ 1.0 & 30.9 $\pm$ 0.7 & 13.2 $\pm$ 0.3 & 2.6 $\pm$ 0.1 & 27.5 $\pm$ 0.4 & 62.5 $\pm$ 0.3 & 56.4 $\pm$ 0.4 \\
	 8 & 72.9 $\pm$ 0.3 & 50.3 $\pm$ 1.4 & 32.2 $\pm$ 2.3 & 61.4 $\pm$ 1.0 & 29.2 $\pm$ 1.0 & 12.9 $\pm$ 0.3 & 2.8 $\pm$ 0.2 & 27.4 $\pm$ 0.4 & 62.6 $\pm$ 0.4 & 56.4 $\pm$ 0.3 \\
	 9 & 72.7 $\pm$ 0.3 & 48.9 $\pm$ 1.9 & 39.4 $\pm$ 2.5 & 62.5 $\pm$ 1.2 & 29.9 $\pm$ 0.7 & 13.1 $\pm$ 0.3 & 2.8 $\pm$ 0.1 & 27.1 $\pm$ 0.4 & 62.2 $\pm$ 0.4 & 55.7 $\pm$ 0.4 \\
   \bottomrule
  \end{tabular}
  }
\end{table}

\subsubsection{DeepAug ERM (EDSR)\cite{hendrycks2021many}} 
\begin{table}[H]
 \centering
 \resizebox{\textwidth}{!}{
 \begin{tabular}{@{}lcccccccccc@{}}
   \toprule
    Run & ImageNet1k & Silhouette & Edge & Sketch & CueConflict & ImageNetStylized & ImageNetA & ImageNetR & DeepAugCAE & DeepAugEDSR \\
   \midrule
   	 1 & 71.8 $\pm$ 0.7 & 52.3 $\pm$ 2.2 & 32.2 $\pm$ 2.9 & 61.5 $\pm$ 1.5 & 31.5 $\pm$ 1.0 & 10.5 $\pm$ 0.4 & 2.0 $\pm$ 0.2 & 25.4 $\pm$ 0.6 & 51.1 $\pm$ 0.9 & 64.0 $\pm$ 0.8 \\
	 2 & 71.8 $\pm$ 0.7 & 51.9 $\pm$ 1.7 & 35.0 $\pm$ 3.7 & 61.6 $\pm$ 1.8 & 33.2 $\pm$ 1.0 & 10.4 $\pm$ 0.5 & 2.0 $\pm$ 0.1 & 25.8 $\pm$ 0.5 & 51.0 $\pm$ 0.8 & 64.1 $\pm$ 0.8 \\
	 3 & 71.9 $\pm$ 0.6 & 51.0 $\pm$ 1.9 & 30.9 $\pm$ 2.2 & 60.3 $\pm$ 1.6 & 31.1 $\pm$ 0.9 & 10.9 $\pm$ 0.5 & 1.9 $\pm$ 0.2 & 25.8 $\pm$ 0.5 & 51.4 $\pm$ 0.9 & 64.1 $\pm$ 0.8 \\
	 4 & 71.9 $\pm$ 0.7 & 52.4 $\pm$ 2.1 & 31.7 $\pm$ 3.0 & 61.6 $\pm$ 1.6 & 32.0 $\pm$ 1.1 & 10.7 $\pm$ 0.5 & 2.1 $\pm$ 0.2 & 25.8 $\pm$ 0.7 & 51.5 $\pm$ 0.8 & 64.2 $\pm$ 0.8 \\
	 5 & 72.1 $\pm$ 0.6 & 52.2 $\pm$ 2.1 & 37.4 $\pm$ 3.6 & 60.7 $\pm$ 1.4 & 33.3 $\pm$ 0.8 & 10.4 $\pm$ 0.5 & 2.1 $\pm$ 0.2 & 25.9 $\pm$ 0.5 & 51.2 $\pm$ 0.9 & 64.5 $\pm$ 0.8 \\
	 6 & 72.1 $\pm$ 0.6 & 51.4 $\pm$ 1.6 & 32.1 $\pm$ 3.5 & 59.7 $\pm$ 1.6 & 32.4 $\pm$ 1.0 & 10.7 $\pm$ 0.4 & 2.1 $\pm$ 0.2 & 25.9 $\pm$ 0.6 & 51.1 $\pm$ 0.8 & 64.2 $\pm$ 0.8 \\
	 7 & 71.7 $\pm$ 0.6 & 51.3 $\pm$ 2.4 & 29.3 $\pm$ 3.0 & 58.6 $\pm$ 1.0 & 33.3 $\pm$ 0.9 & 10.4 $\pm$ 0.4 & 2.1 $\pm$ 0.2 & 25.9 $\pm$ 0.5 & 51.1 $\pm$ 0.9 & 64.0 $\pm$ 0.8 \\
	 8 & 71.9 $\pm$ 0.7 & 49.9 $\pm$ 2.6 & 29.5 $\pm$ 3.2 & 59.5 $\pm$ 1.6 & 31.8 $\pm$ 0.9 & 10.7 $\pm$ 0.5 & 1.8 $\pm$ 0.2 & 25.1 $\pm$ 0.6 & 50.7 $\pm$ 0.8 & 64.2 $\pm$ 0.8 \\
	 9 & 71.8 $\pm$ 0.7 & 51.7 $\pm$ 2.0 & 27.5 $\pm$ 3.0 & 59.0 $\pm$ 1.4 & 31.8 $\pm$ 0.9 & 11.0 $\pm$ 0.5 & 1.9 $\pm$ 0.1 & 25.8 $\pm$ 0.7 & 51.2 $\pm$ 0.8 & 64.1 $\pm$ 0.8 \\
	 10 & 72.0 $\pm$ 0.7 & 49.8 $\pm$ 2.2 & 34.8 $\pm$ 3.2 & 61.1 $\pm$ 1.7 & 32.3 $\pm$ 0.9 & 10.6 $\pm$ 0.5 & 2.3 $\pm$ 0.2 & 26.1 $\pm$ 0.4 & 51.1 $\pm$ 0.8 & 64.2 $\pm$ 0.8 \\
   \bottomrule
  \end{tabular}
  }
\end{table}

\subsubsection{Stylized ImageNet \cite{geirhos2018imagenet}}
\begin{table}[H]
 \centering
 \resizebox{\textwidth}{!}{
 \begin{tabular}{@{}lcccccccccc@{}}
   \toprule
    Run & ImageNet1k & Silhouette & Edge & Sketch & CueConflict & ImageNetStylized & ImageNetA & ImageNetR & DeepAugCAE & DeepAugEDSR \\
   \midrule
   	 1 & 56.2 $\pm$ 0.4 & 47.7 $\pm$ 1.3 & 61.8 $\pm$ 1.9 & 71.1 $\pm$ 0.7 & 54.5 $\pm$ 1.0 & 52.6 $\pm$ 0.5 & 0.8 $\pm$ 0.1 & 25.5 $\pm$ 0.4 & 39.8 $\pm$ 0.4 & 40.8 $\pm$ 0.4 \\
	 2 & 55.3 $\pm$ 0.5 & 46.7 $\pm$ 1.7 & 59.3 $\pm$ 2.8 & 70.1 $\pm$ 0.8 & 54.0 $\pm$ 1.0 & 52.8 $\pm$ 0.5 & 0.7 $\pm$ 0.1 & 24.7 $\pm$ 0.4 & 38.5 $\pm$ 0.6 & 40.0 $\pm$ 0.5 \\
	 3 & 54.6 $\pm$ 0.5 & 43.7 $\pm$ 2.9 & 61.6 $\pm$ 2.6 & 69.5 $\pm$ 1.0 & 55.7 $\pm$ 1.1 & 52.1 $\pm$ 0.7 & 0.7 $\pm$ 0.1 & 24.7 $\pm$ 0.4 & 38.4 $\pm$ 0.5 & 39.6 $\pm$ 0.4 \\
	 4 & 55.4 $\pm$ 0.5 & 47.8 $\pm$ 1.7 & 61.4 $\pm$ 2.8 & 69.6 $\pm$ 1.1 & 54.0 $\pm$ 1.2 & 52.4 $\pm$ 0.6 & 0.8 $\pm$ 0.1 & 25.1 $\pm$ 0.4 & 38.8 $\pm$ 0.5 & 40.2 $\pm$ 0.4 \\
	 5 & 55.5 $\pm$ 0.5 & 46.7 $\pm$ 1.4 & 58.4 $\pm$ 2.9 & 71.1 $\pm$ 0.8 & 52.5 $\pm$ 1.4 & 53.0 $\pm$ 0.5 & 0.7 $\pm$ 0.1 & 25.3 $\pm$ 0.4 & 38.7 $\pm$ 0.4 & 40.3 $\pm$ 0.4 \\
	 6 & 54.8 $\pm$ 0.5 & 50.8 $\pm$ 1.4 & 59.9 $\pm$ 2.7 & 69.1 $\pm$ 0.9 & 54.0 $\pm$ 1.4 & 52.1 $\pm$ 0.7 & 0.7 $\pm$ 0.1 & 24.6 $\pm$ 0.4 & 38.3 $\pm$ 0.6 & 39.5 $\pm$ 0.5 \\
	 7 & 55.2 $\pm$ 0.5 & 45.1 $\pm$ 2.2 & 56.0 $\pm$ 2.5 & 70.6 $\pm$ 1.0 & 54.1 $\pm$ 0.8 & 52.3 $\pm$ 0.6 & 0.8 $\pm$ 0.1 & 24.7 $\pm$ 0.5 & 38.3 $\pm$ 0.6 & 40.0 $\pm$ 0.5 \\
	 8 & 55.9 $\pm$ 0.4 & 46.6 $\pm$ 1.8 & 56.3 $\pm$ 2.9 & 71.0 $\pm$ 0.6 & 54.2 $\pm$ 1.3 & 52.6 $\pm$ 0.5 & 0.8 $\pm$ 0.1 & 25.5 $\pm$ 0.5 & 39.1 $\pm$ 0.5 & 40.6 $\pm$ 0.5 \\
	 9 & 55.3 $\pm$ 0.6 & 49.2 $\pm$ 1.8 & 62.0 $\pm$ 3.1 & 68.7 $\pm$ 1.0 & 55.7 $\pm$ 1.2 & 52.4 $\pm$ 0.7 & 0.7 $\pm$ 0.1 & 25.1 $\pm$ 0.3 & 38.8 $\pm$ 0.6 & 40.1 $\pm$ 0.5 \\
	 10 & 54.9 $\pm$ 0.4 & 48.3 $\pm$ 2.0 & 58.2 $\pm$ 3.5 & 70.7 $\pm$ 0.9 & 53.3 $\pm$ 1.4 & 52.0 $\pm$ 0.5 & 0.7 $\pm$ 0.1 & 24.7 $\pm$ 0.3 & 38.8 $\pm$ 0.4 & 40.0 $\pm$ 0.4 \\
   \bottomrule
  \end{tabular}
  }
\end{table}

\subsubsection{InfoDrop \cite{shi2020informative}} 
\begin{table}[H]
 \centering
 \resizebox{\textwidth}{!}{
 \begin{tabular}{@{}lcccccccccc@{}}
   \toprule
    Run & ImageNet1k & Silhouette & Edge & Sketch & CueConflict & ImageNetStylized & ImageNetA & ImageNetR & DeepAugCAE & DeepAugEDSR \\
   \midrule
   	 1 & 73.2 $\pm$ 0.4 & 43.9 $\pm$ 1.7 & 22.4 $\pm$ 3.0 & 57.9 $\pm$ 1.1 & 23.1 $\pm$ 0.7 & 7.8 $\pm$ 0.3 & 2.3 $\pm$ 0.2 & 23.2 $\pm$ 0.5 & 44.3 $\pm$ 0.5 & 48.1 $\pm$ 0.4 \\
	 2 & 72.1 $\pm$ 0.7 & 47.4 $\pm$ 2.0 & 22.9 $\pm$ 3.1 & 57.1 $\pm$ 1.7 & 21.9 $\pm$ 0.7 & 7.5 $\pm$ 0.4 & 2.0 $\pm$ 0.2 & 22.3 $\pm$ 0.6 & 43.0 $\pm$ 0.9 & 48.0 $\pm$ 0.7 \\
	 3 & 72.8 $\pm$ 0.3 & 47.1 $\pm$ 2.0 & 22.1 $\pm$ 1.5 & 55.7 $\pm$ 1.1 & 22.4 $\pm$ 0.5 & 7.6 $\pm$ 0.3 & 2.1 $\pm$ 0.2 & 22.6 $\pm$ 0.3 & 44.0 $\pm$ 0.5 & 48.6 $\pm$ 0.3 \\
	 4 & 73.0 $\pm$ 0.4 & 48.0 $\pm$ 2.4 & 15.5 $\pm$ 1.4 & 55.9 $\pm$ 1.2 & 22.5 $\pm$ 0.6 & 7.6 $\pm$ 0.2 & 2.1 $\pm$ 0.1 & 22.6 $\pm$ 0.4 & 44.0 $\pm$ 0.5 & 48.9 $\pm$ 0.4 \\
	 5 & 72.8 $\pm$ 0.3 & 46.3 $\pm$ 1.4 & 18.2 $\pm$ 1.9 & 56.4 $\pm$ 0.9 & 22.9 $\pm$ 0.7 & 7.6 $\pm$ 0.3 & 2.2 $\pm$ 0.2 & 22.6 $\pm$ 0.5 & 43.7 $\pm$ 0.5 & 48.2 $\pm$ 0.4 \\
	 6 & 73.1 $\pm$ 0.4 & 46.5 $\pm$ 1.9 & 16.8 $\pm$ 2.8 & 58.5 $\pm$ 1.5 & 22.9 $\pm$ 0.6 & 7.6 $\pm$ 0.3 & 2.2 $\pm$ 0.2 & 22.9 $\pm$ 0.4 & 44.2 $\pm$ 0.5 & 49.0 $\pm$ 0.5 \\
	 7 & 72.9 $\pm$ 0.3 & 49.7 $\pm$ 1.2 & 13.2 $\pm$ 1.4 & 55.1 $\pm$ 1.2 & 23.7 $\pm$ 0.6 & 7.8 $\pm$ 0.2 & 2.2 $\pm$ 0.2 & 22.5 $\pm$ 0.4 & 44.0 $\pm$ 0.5 & 48.6 $\pm$ 0.5 \\
	 8 & 73.1 $\pm$ 0.4 & 47.4 $\pm$ 1.6 & 23.8 $\pm$ 2.2 & 55.8 $\pm$ 0.9 & 22.8 $\pm$ 0.8 & 7.9 $\pm$ 0.2 & 2.1 $\pm$ 0.2 & 22.7 $\pm$ 0.4 & 44.0 $\pm$ 0.6 & 48.2 $\pm$ 0.5 \\
	 9 & 72.8 $\pm$ 0.3 & 50.1 $\pm$ 1.6 & 13.1 $\pm$ 1.3 & 55.2 $\pm$ 1.3 & 23.7 $\pm$ 0.6 & 7.8 $\pm$ 0.2 & 2.2 $\pm$ 0.1 & 22.6 $\pm$ 0.4 & 44.0 $\pm$ 0.5 & 48.6 $\pm$ 0.5 \\
	 10 & 72.8 $\pm$ 0.3 & 43.4 $\pm$ 1.7 & 21.5 $\pm$ 2.2 & 58.0 $\pm$ 1.0 & 23.3 $\pm$ 0.8 & 8.2 $\pm$ 0.3 & 2.3 $\pm$ 0.1 & 22.8 $\pm$ 0.5 & 44.2 $\pm$ 0.4 & 47.8 $\pm$ 0.3 \\
   \bottomrule
  \end{tabular}
  }
\end{table}

\subsubsection{SagNet \cite{nam2021reducing}} 
\begin{table}[H]
 \centering
 \resizebox{\textwidth}{!}{
 \begin{tabular}{@{}lcccccccccc@{}}
   \toprule
    Run & ImageNet1k & Silhouette & Edge & Sketch & CueConflict & ImageNetStylized & ImageNetA & ImageNetR & DeepAugCAE & DeepAugEDSR \\
   \midrule
   	 1 & 74.8 $\pm$ 0.3 & 45.2 $\pm$ 1.6 & 29.8 $\pm$ 1.9 & 59.6 $\pm$ 0.9 & 19.8 $\pm$ 0.4 & 6.4 $\pm$ 0.2 & 1.8 $\pm$ 0.2 & 22.8 $\pm$ 0.5 & 44.1 $\pm$ 0.5 & 48.1 $\pm$ 0.3 \\
	 2 & 73.7 $\pm$ 0.2 & 44.9 $\pm$ 0.7 & 25.4 $\pm$ 1.1 & 60.4 $\pm$ 1.2 & 20.1 $\pm$ 0.4 & 6.1 $\pm$ 0.2 & 1.7 $\pm$ 0.1 & 22.4 $\pm$ 0.5 & 42.9 $\pm$ 0.5 & 47.2 $\pm$ 0.3 \\
	 3 & 73.7 $\pm$ 0.1 & 44.0 $\pm$ 1.1 & 28.8 $\pm$ 2.1 & 60.3 $\pm$ 1.1 & 20.1 $\pm$ 0.4 & 6.3 $\pm$ 0.2 & 1.7 $\pm$ 0.1 & 22.4 $\pm$ 0.4 & 43.1 $\pm$ 0.5 & 47.1 $\pm$ 0.3 \\
	 4 & 73.6 $\pm$ 0.2 & 45.0 $\pm$ 1.0 & 27.0 $\pm$ 2.3 & 59.9 $\pm$ 0.9 & 19.3 $\pm$ 0.4 & 6.2 $\pm$ 0.2 & 1.5 $\pm$ 0.1 & 22.4 $\pm$ 0.4 & 43.0 $\pm$ 0.5 & 47.1 $\pm$ 0.4 \\
	 5 & 73.6 $\pm$ 0.1 & 43.4 $\pm$ 1.0 & 24.2 $\pm$ 1.8 & 60.1 $\pm$ 1.2 & 19.4 $\pm$ 0.4 & 6.2 $\pm$ 0.2 & 1.7 $\pm$ 0.1 & 22.4 $\pm$ 0.5 & 43.1 $\pm$ 0.4 & 47.2 $\pm$ 0.3 \\
	 6 & 74.7 $\pm$ 0.3 & 45.4 $\pm$ 1.1 & 27.9 $\pm$ 1.8 & 59.7 $\pm$ 1.0 & 19.9 $\pm$ 0.5 & 6.4 $\pm$ 0.2 & 1.8 $\pm$ 0.1 & 22.8 $\pm$ 0.4 & 44.0 $\pm$ 0.5 & 48.1 $\pm$ 0.4 \\
	 7 & 73.6 $\pm$ 0.1 & 45.9 $\pm$ 1.2 & 27.1 $\pm$ 2.5 & 61.1 $\pm$ 1.2 & 19.6 $\pm$ 0.4 & 6.1 $\pm$ 0.2 & 1.8 $\pm$ 0.2 & 22.5 $\pm$ 0.5 & 43.0 $\pm$ 0.5 & 47.0 $\pm$ 0.3 \\
	 8 & 73.6 $\pm$ 0.2 & 44.5 $\pm$ 1.2 & 27.2 $\pm$ 1.9 & 60.9 $\pm$ 1.0 & 19.4 $\pm$ 0.4 & 6.1 $\pm$ 0.2 & 1.6 $\pm$ 0.1 & 22.4 $\pm$ 0.4 & 43.0 $\pm$ 0.6 & 47.2 $\pm$ 0.4 \\
	 9 & 73.6 $\pm$ 0.1 & 43.5 $\pm$ 1.2 & 27.2 $\pm$ 1.9 & 61.0 $\pm$ 1.1 & 20.1 $\pm$ 0.4 & 6.2 $\pm$ 0.2 & 1.5 $\pm$ 0.1 & 22.5 $\pm$ 0.4 & 43.1 $\pm$ 0.5 & 47.2 $\pm$ 0.3 \\
	 10 & 73.6 $\pm$ 0.2 & 44.6 $\pm$ 1.2 & 26.4 $\pm$ 1.7 & 60.9 $\pm$ 1.3 & 19.4 $\pm$ 0.4 & 6.2 $\pm$ 0.2 & 1.7 $\pm$ 0.2 & 22.4 $\pm$ 0.5 & 42.9 $\pm$ 0.6 & 47.1 $\pm$ 0.4 \\
   \bottomrule
  \end{tabular}
  }
\end{table}

\subsubsection{pAdaIN \cite{nuriel2021permuted}} 
\begin{table}[H]
 \centering
 \resizebox{\textwidth}{!}{
 \begin{tabular}{@{}lcccccccccc@{}}
   \toprule
    Run & ImageNet1k & Silhouette & Edge & Sketch & CueConflict & ImageNetStylized & ImageNetA & ImageNetR & DeepAugCAE & DeepAugEDSR \\
   \midrule
   	 1 & 72.9 $\pm$ 0.1 & 45.0 $\pm$ 0.9 & 18.8 $\pm$ 1.9 & 55.4 $\pm$ 0.7 & 21.6 $\pm$ 0.4 & 8.3 $\pm$ 0.1 & 1.4 $\pm$ 0.1 & 21.2 $\pm$ 0.2 & 42.7 $\pm$ 0.2 & 49.0 $\pm$ 0.2 \\
	 2 & 73.0 $\pm$ 0.1 & 47.0 $\pm$ 0.8 & 13.3 $\pm$ 0.9 & 56.1 $\pm$ 0.9 & 21.7 $\pm$ 0.4 & 8.3 $\pm$ 0.1 & 1.4 $\pm$ 0.1 & 21.5 $\pm$ 0.2 & 42.7 $\pm$ 0.2 & 48.1 $\pm$ 0.2 \\
	 3 & 73.0 $\pm$ 0.1 & 44.5 $\pm$ 1.2 & 25.0 $\pm$ 1.0 & 57.0 $\pm$ 0.7 & 20.5 $\pm$ 0.4 & 8.0 $\pm$ 0.1 & 1.5 $\pm$ 0.1 & 21.2 $\pm$ 0.2 & 42.8 $\pm$ 0.2 & 48.6 $\pm$ 0.2 \\
	 4 & 73.0 $\pm$ 0.1 & 44.4 $\pm$ 1.2 & 21.7 $\pm$ 3.4 & 56.0 $\pm$ 1.1 & 21.7 $\pm$ 0.4 & 8.3 $\pm$ 0.1 & 1.5 $\pm$ 0.1 & 21.4 $\pm$ 0.3 & 42.7 $\pm$ 0.2 & 48.9 $\pm$ 0.2 \\
	 5 & 73.0 $\pm$ 0.1 & 40.7 $\pm$ 1.4 & 18.6 $\pm$ 1.3 & 56.6 $\pm$ 0.7 & 20.3 $\pm$ 0.3 & 7.8 $\pm$ 0.1 & 1.5 $\pm$ 0.1 & 21.7 $\pm$ 0.2 & 42.4 $\pm$ 0.2 & 48.3 $\pm$ 0.2 \\
	 6 & 73.0 $\pm$ 0.1 & 46.7 $\pm$ 1.0 & 23.1 $\pm$ 1.1 & 57.4 $\pm$ 0.7 & 22.1 $\pm$ 0.4 & 8.1 $\pm$ 0.1 & 1.4 $\pm$ 0.1 & 21.4 $\pm$ 0.2 & 42.8 $\pm$ 0.2 & 48.5 $\pm$ 0.2 \\
	 7 & 73.1 $\pm$ 0.1 & 43.7 $\pm$ 1.0 & 22.0 $\pm$ 0.9 & 55.9 $\pm$ 0.6 & 21.3 $\pm$ 0.5 & 8.1 $\pm$ 0.1 & 1.5 $\pm$ 0.1 & 21.8 $\pm$ 0.2 & 42.7 $\pm$ 0.2 & 48.6 $\pm$ 0.2 \\
	 8 & 73.2 $\pm$ 0.1 & 45.8 $\pm$ 1.1 & 19.7 $\pm$ 0.8 & 56.9 $\pm$ 0.7 & 20.7 $\pm$ 0.4 & 8.1 $\pm$ 0.1 & 1.5 $\pm$ 0.1 & 21.3 $\pm$ 0.2 & 42.7 $\pm$ 0.2 & 48.6 $\pm$ 0.2 \\
	 9 & 73.1 $\pm$ 0.1 & 45.0 $\pm$ 2.2 & 23.2 $\pm$ 4.2 & 56.4 $\pm$ 0.7 & 22.0 $\pm$ 0.3 & 8.2 $\pm$ 0.1 & 1.5 $\pm$ 0.1 & 21.7 $\pm$ 0.2 & 42.8 $\pm$ 0.2 & 48.7 $\pm$ 0.2 \\
	 10 & 73.1 $\pm$ 0.1 & 40.8 $\pm$ 1.2 & 23.4 $\pm$ 1.3 & 56.6 $\pm$ 0.5 & 21.3 $\pm$ 0.4 & 7.9 $\pm$ 0.1 & 1.4 $\pm$ 0.1 & 21.3 $\pm$ 0.2 & 42.3 $\pm$ 0.1 & 48.3 $\pm$ 0.1 \\
   \bottomrule
  \end{tabular}
  }
\end{table}

\subsection{Model selection: best validation score}

\subsubsection{ERM \cite{vapnik1999overview}} 
\begin{table}[H]
 \centering
 \resizebox{\textwidth}{!}{
 \begin{tabular}{@{}lcccccccccc@{}}
   \toprule
    Run & ImageNet1k & Silhouette & Edge & Sketch & CueConflict & ImageNetStylized & ImageNetA & ImageNetR & DeepAugCAE & DeepAugEDSR \\
   \midrule
   	 1 & 73.6 & 48.1 & 23.1 & 56.2 & 22.0 & 7.8 & 2.0 & 22.2 & 44.1 & 47.6 \\
	 2 & 73.5 & 42.5 & 26.2 & 56.5 & 23.0 & 7.9 & 1.7 & 23.5 & 44.5 & 48.3 \\
	 3 & 74.0 & 46.9 & 20.0 & 54.3 & 22.0 & 8.0 & 1.9 & 22.6 & 44.9 & 48.6 \\
	 4 & 73.7 & 50.0 & 18.1 & 56.0 & 21.9 & 8.0 & 2.4 & 22.8 & 44.1 & 47.6 \\
	 5 & 73.9 & 46.3 & 25.6 & 56.4 & 21.3 & 7.9 & 2.2 & 22.9 & 44.6 & 49.1 \\
	 6 & 73.7 & 48.1 & 20.6 & 56.5 & 22.7 & 7.8 & 2.0 & 23.2 & 44.3 & 48.0 \\
	 7 & 73.8 & 48.1 & 22.5 & 57.6 & 20.9 & 8.0 & 2.3 & 22.5 & 44.0 & 48.3 \\
	 8 & 73.8 & 50.0 & 20.0 & 56.0 & 24.2 & 7.8 & 2.0 & 23.2 & 44.7 & 48.1 \\
	 9 & 73.9 & 48.8 & 21.3 & 54.3 & 21.7 & 8.0 & 2.1 & 22.6 & 44.7 & 48.8 \\
	 10 & 73.9 & 44.4 & 28.7 & 55.9 & 22.0 & 7.5 & 2.0 & 22.6 & 44.4 & 48.4 \\
   \bottomrule
  \end{tabular}
  }
\end{table}

\subsubsection{Debiased \cite{li2020shape}} 
\begin{table}[H]
 \centering
 \resizebox{\textwidth}{!}{
 \begin{tabular}{@{}lcccccccccc@{}}
   \toprule
    Run & ImageNet1k & Silhouette & Edge & Sketch & CueConflict & ImageNetStylized & ImageNetA & ImageNetR & DeepAugCAE & DeepAugEDSR \\
   \midrule
   	 1 & 74.4 & 51.9 & 33.1 & 60.1 & 28.8 & 15.9 & 2.7 & 27.4 & 49.5 & 51.7 \\
	 2 & 74.4 & 48.8 & 21.3 & 61.0 & 27.8 & 15.9 & 2.4 & 27.2 & 49.6 & 51.2 \\
	 3 & 74.6 & 46.9 & 38.7 & 60.1 & 29.1 & 16.2 & 2.5 & 27.5 & 49.3 & 51.5 \\
	 4 & 74.6 & 45.0 & 27.5 & 60.9 & 30.5 & 16.4 & 2.8 & 27.2 & 49.8 & 50.8 \\
	 5 & 74.2 & 48.1 & 35.6 & 59.5 & 29.0 & 15.6 & 2.5 & 26.8 & 49.6 & 51.2 \\
	 6 & 74.7 & 48.8 & 26.9 & 60.0 & 29.1 & 16.2 & 3.2 & 27.8 & 50.1 & 51.8 \\
	 7 & 74.5 & 43.8 & 33.1 & 58.4 & 27.4 & 16.0 & 2.7 & 27.7 & 49.6 & 51.4 \\
	 8 & 74.4 & 51.9 & 33.8 & 62.7 & 28.4 & 16.6 & 2.7 & 27.1 & 49.5 & 51.4 \\
	 9 & 74.3 & 51.2 & 30.6 & 61.9 & 30.9 & 16.2 & 2.8 & 28.0 & 49.7 & 51.1 \\
	 10 & 74.5 & 50.6 & 27.5 & 60.1 & 28.3 & 15.5 & 2.8 & 27.1 & 49.8 & 50.9 \\
   \bottomrule
  \end{tabular}
  }
\end{table}

\subsubsection{DeepAug ERM (CAE) \cite{hendrycks2021many}} 
\begin{table}[H]
 \centering
 \resizebox{\textwidth}{!}{
 \begin{tabular}{@{}lcccccccccc@{}}
   \toprule
    Run & ImageNet1k & Silhouette & Edge & Sketch & CueConflict & ImageNetStylized & ImageNetA & ImageNetR & DeepAugCAE & DeepAugEDSR \\
   \midrule
   	 1 & 73.9 & 52.5 & 38.1 & 62.0 & 31.1 & 13.3 & 2.6 & 29.3 & 63.5 & 57.1 \\
	 2 & 73.7 & 55.6 & 38.7 & 63.7 & 31.4 & 13.1 & 2.4 & 28.0 & 63.0 & 56.0 \\
	 3 & 73.3 & 50.6 & 33.8 & 64.5 & 29.5 & 13.7 & 2.5 & 27.6 & 63.2 & 56.4 \\
	 4 & 73.7 & 48.8 & 30.6 & 65.9 & 30.3 & 13.1 & 2.5 & 28.9 & 63.2 & 56.2 \\
	 5 & 73.6 & 50.6 & 27.5 & 63.7 & 30.1 & 13.3 & 2.7 & 27.8 & 63.2 & 56.9 \\
	 6 & 73.8 & 53.8 & 39.4 & 66.5 & 31.4 & 13.2 & 2.8 & 28.8 & 63.3 & 57.1 \\
	 7 & 73.6 & 52.5 & 42.5 & 65.6 & 31.2 & 13.7 & 2.9 & 28.4 & 63.3 & 57.3 \\
	 8 & 73.6 & 55.0 & 38.1 & 62.6 & 30.8 & 13.1 & 2.5 & 28.0 & 63.2 & 57.1 \\
	 9 & 73.5 & 48.1 & 40.0 & 62.6 & 29.6 & 13.3 & 2.6 & 28.1 & 62.9 & 56.2 \\
   \bottomrule
  \end{tabular}
  }
\end{table}

\subsubsection{DeepAug ERM (EDSR) \cite{hendrycks2021many}} 
\begin{table}[H]
 \centering
 \resizebox{\textwidth}{!}{
 \begin{tabular}{@{}lcccccccccc@{}}
   \toprule
    Run & ImageNet1k & Silhouette & Edge & Sketch & CueConflict & ImageNetStylized & ImageNetA & ImageNetR & DeepAugCAE & DeepAugEDSR \\
   \midrule
   	 1 & 72.7 & 51.2 & 34.4 & 63.5 & 31.8 & 10.9 & 2.0 & 26.5 & 52.3 & 65.0 \\
	 2 & 72.7 & 52.5 & 30.0 & 59.8 & 32.9 & 10.6 & 2.0 & 26.4 & 52.3 & 65.1 \\
	 3 & 72.9 & 53.8 & 28.7 & 60.1 & 29.5 & 11.3 & 1.9 & 26.8 & 52.6 & 65.3 \\
	 4 & 73.0 & 50.0 & 28.7 & 61.3 & 32.0 & 11.0 & 1.9 & 26.2 & 52.6 & 65.2 \\
	 5 & 73.2 & 50.6 & 39.4 & 62.6 & 35.2 & 10.8 & 2.3 & 27.5 & 52.6 & 65.3 \\
	 6 & 73.2 & 51.2 & 23.7 & 62.7 & 31.6 & 11.3 & 1.9 & 26.6 & 52.5 & 65.1 \\
	 7 & 72.6 & 51.2 & 31.2 & 59.9 & 33.1 & 10.7 & 2.2 & 26.8 & 52.3 & 64.8 \\
	 8 & 72.7 & 53.1 & 30.0 & 58.6 & 31.5 & 11.0 & 1.8 & 25.5 & 51.8 & 65.2 \\
	 9 & 72.6 & 51.9 & 33.1 & 61.1 & 32.8 & 11.6 & 1.9 & 26.1 & 52.3 & 65.0 \\
	 10 & 72.9 & 50.0 & 37.5 & 61.5 & 33.4 & 11.0 & 2.4 & 26.7 & 52.0 & 65.0 \\
   \bottomrule
  \end{tabular}
  }
\end{table}

\subsubsection{Stylized ImageNet \cite{geirhos2018imagenet}}
\begin{table}[H]
 \centering
 \resizebox{\textwidth}{!}{
 \begin{tabular}{@{}lcccccccccc@{}}
   \toprule
    Run & ImageNet1k & Silhouette & Edge & Sketch & CueConflict & ImageNetStylized & ImageNetA & ImageNetR & DeepAugCAE & DeepAugEDSR \\
   \midrule
   	 1 & 56.7 & 47.5 & 63.1 & 71.2 & 53.4 & 53.2 & 0.9 & 25.5 & 40.1 & 40.9 \\
	 2 & 56.1 & 47.5 & 59.4 & 70.2 & 55.2 & 53.5 & 0.8 & 24.9 & 39.0 & 40.4 \\
	 3 & 55.3 & 41.2 & 60.0 & 69.1 & 55.3 & 53.1 & 0.7 & 24.9 & 39.0 & 40.1 \\
	 4 & 55.9 & 45.6 & 58.1 & 69.0 & 52.5 & 53.3 & 0.9 & 25.0 & 39.0 & 40.3 \\
	 5 & 56.1 & 45.0 & 56.2 & 72.0 & 51.4 & 53.6 & 0.8 & 25.4 & 38.8 & 40.5 \\
	 6 & 55.7 & 50.6 & 61.3 & 70.1 & 54.5 & 53.1 & 0.7 & 24.8 & 39.0 & 39.9 \\
	 7 & 55.7 & 45.0 & 58.1 & 71.6 & 53.8 & 53.1 & 0.8 & 24.8 & 38.4 & 40.0 \\
	 8 & 56.3 & 47.5 & 55.6 & 71.0 & 54.1 & 53.3 & 0.8 & 25.1 & 39.5 & 40.9 \\
	 9 & 56.0 & 48.8 & 60.0 & 67.5 & 54.9 & 53.4 & 0.6 & 25.1 & 39.2 & 40.5 \\
	 10 & 55.2 & 50.0 & 52.5 & 69.7 & 51.6 & 52.8 & 0.7 & 24.5 & 38.8 & 40.0 \\
   \bottomrule
  \end{tabular}
  }
\end{table}

\subsubsection{InfoDrop \cite{shi2020informative}} 
\begin{table}[H]
 \centering
 \resizebox{\textwidth}{!}{
 \begin{tabular}{@{}lcccccccccc@{}}
   \toprule
    Run & ImageNet1k & Silhouette & Edge & Sketch & CueConflict & ImageNetStylized & ImageNetA & ImageNetR & DeepAugCAE & DeepAugEDSR \\
   \midrule
   	 1 & 73.7 & 45.0 & 25.0 & 56.5 & 23.3 & 7.8 & 2.5 & 23.2 & 44.4 & 48.4 \\
	 2 & 73.1 & 49.4 & 20.0 & 58.5 & 22.4 & 7.9 & 2.2 & 22.8 & 43.9 & 48.7 \\
	 3 & 73.2 & 48.1 & 21.3 & 57.6 & 22.4 & 7.6 & 2.2 & 22.4 & 44.4 & 48.5 \\
	 4 & 73.4 & 50.0 & 16.2 & 57.4 & 22.6 & 7.6 & 2.3 & 22.6 & 44.6 & 49.5 \\
	 5 & 73.1 & 50.0 & 20.0 & 55.8 & 22.7 & 8.0 & 2.2 & 23.0 & 44.8 & 48.6 \\
	 6 & 73.7 & 45.6 & 16.2 & 60.5 & 21.8 & 7.6 & 2.3 & 23.0 & 44.5 & 49.3 \\
	 7 & 73.3 & 49.4 & 12.5 & 54.3 & 23.8 & 7.8 & 2.0 & 22.2 & 44.4 & 48.7 \\
	 8 & 73.6 & 47.5 & 23.7 & 56.0 & 23.4 & 7.9 & 2.2 & 22.5 & 44.3 & 48.2 \\
	 9 & 73.3 & 50.6 & 13.1 & 53.9 & 23.2 & 7.8 & 2.2 & 22.2 & 44.3 & 48.7 \\
	 10 & 73.1 & 41.9 & 21.9 & 56.6 & 22.5 & 8.6 & 2.2 & 22.7 & 44.4 & 47.7 \\
   \bottomrule
  \end{tabular}
  }
\end{table}

\subsubsection{SagNet \cite{nam2021reducing}} 
\begin{table}[H]
 \centering
 \resizebox{\textwidth}{!}{
 \begin{tabular}{@{}lcccccccccc@{}}
   \toprule
    Run & ImageNet1k & Silhouette & Edge & Sketch & CueConflict & ImageNetStylized & ImageNetA & ImageNetR & DeepAugCAE & DeepAugEDSR \\
   \midrule
   	 1 & 75.2 & 43.8 & 29.4 & 59.2 & 19.8 & 6.1 & 1.6 & 22.3 & 44.4 & 48.2 \\
	 2 & 73.9 & 44.4 & 24.4 & 58.4 & 19.8 & 6.4 & 1.8 & 22.1 & 43.2 & 47.2 \\
	 3 & 73.9 & 41.9 & 28.1 & 59.6 & 20.4 & 6.1 & 1.3 & 21.4 & 43.5 & 47.3 \\
	 4 & 73.9 & 45.0 & 23.1 & 59.2 & 19.8 & 6.2 & 1.4 & 21.7 & 44.0 & 47.6 \\
	 5 & 73.9 & 41.9 & 23.7 & 58.7 & 19.7 & 6.2 & 1.5 & 21.7 & 43.8 & 47.3 \\
	 6 & 75.1 & 45.6 & 25.0 & 58.0 & 20.3 & 6.3 & 1.8 & 22.2 & 44.2 & 48.8 \\
	 7 & 73.9 & 45.0 & 26.2 & 62.0 & 19.7 & 6.2 & 1.9 & 22.5 & 43.3 & 47.5 \\
	 8 & 74.0 & 42.5 & 24.4 & 59.2 & 20.0 & 6.0 & 1.2 & 21.3 & 44.1 & 47.8 \\
	 9 & 73.9 & 41.2 & 24.4 & 59.5 & 19.9 & 6.1 & 1.4 & 21.7 & 43.9 & 47.9 \\
	 10 & 74.0 & 41.9 & 24.4 & 58.2 & 20.2 & 6.3 & 1.5 & 21.1 & 43.8 & 47.5 \\
   \bottomrule
  \end{tabular}
  }
\end{table}

\subsubsection{pAdaIN \cite{nuriel2021permuted}} 
\begin{table}[H]
 \centering
 \resizebox{\textwidth}{!}{
 \begin{tabular}{@{}lcccccccccc@{}}
   \toprule
    Run & ImageNet1k & Silhouette & Edge & Sketch & CueConflict & ImageNetStylized & ImageNetA & ImageNetR & DeepAugCAE & DeepAugEDSR \\
   \midrule
   	 1 & 73.0 & 44.4 & 21.3 & 56.0 & 21.3 & 8.3 & 1.4 & 21.3 & 43.0 & 49.3 \\
	 2 & 73.2 & 48.1 & 13.1 & 56.7 & 21.6 & 8.1 & 1.5 & 21.7 & 42.8 & 48.6 \\
	 3 & 73.1 & 41.2 & 23.7 & 56.2 & 20.5 & 8.1 & 1.3 & 21.1 & 42.7 & 48.6 \\
	 4 & 73.0 & 44.4 & 21.3 & 56.0 & 21.3 & 8.3 & 1.4 & 21.3 & 43.0 & 49.3 \\
	 5 & 73.2 & 42.5 & 18.8 & 57.9 & 20.9 & 8.1 & 1.7 & 22.0 & 42.5 & 48.3 \\
	 6 & 73.1 & 46.3 & 24.4 & 57.1 & 22.3 & 8.2 & 1.4 & 21.3 & 43.2 & 48.7 \\
	 7 & 73.3 & 43.1 & 22.5 & 55.8 & 21.3 & 8.1 & 1.6 & 22.1 & 42.6 & 48.6 \\
	 8 & 73.3 & 45.6 & 20.0 & 58.0 & 21.2 & 8.0 & 1.4 & 21.2 & 42.9 & 48.5 \\
	 9 & 73.2 & 45.6 & 24.4 & 56.9 & 22.0 & 8.2 & 1.6 & 21.6 & 42.9 & 48.9 \\
	 10 & 73.3 & 40.0 & 21.3 & 57.1 & 21.7 & 8.0 & 1.4 & 21.2 & 42.5 & 48.6 \\
   \bottomrule
  \end{tabular}
  }
\end{table}

